%% file: main.tex
\documentclass{article}

\usepackage[final]{neurips_2025}

\usepackage[utf8]{inputenc} 
\usepackage[T1]{fontenc}    
\usepackage{hyperref}       
\usepackage{url}            
\usepackage{booktabs}       
\usepackage{amsfonts}       
\usepackage{nicefrac}       
\usepackage{microtype}      
\usepackage{xcolor}         
\usepackage{algorithm}
\usepackage{algpseudocode}
\usepackage{graphicx} 
\usepackage{subcaption}
\usepackage{array}
\usepackage{caption}
\usepackage{wrapfig}
\usepackage{amsmath}
\usepackage{tcolorbox}
\usepackage[capitalize,noabbrev]{cleveref}
\usepackage{tikz}
\usepackage{float}
\usepackage[toc,page,header]{appendix}

\usetikzlibrary{arrows.meta, positioning}

\newcommand\norm[1]{\left\lVert#1\right\rVert}

\input{defs}

\title{Toward Artificial Palpation: Representation Learning of Touch on Soft Bodies}

\author{%
  Zohar Rimon\thanks{Correspondence to zohar.rimon@campus.technion.ac.il. Code and data are available at \href{https://zoharri.github.io/artificial-palpation/}{\texttt{zoharri.github.io/artificial-palpation}}} \\
  \And
  Elisei Shafer \\
  \And
  Tal Tepper \\
  \And
  Efrat Shimron \\
  \And
  Aviv Tamar \\
}

\begin{document}

\vspace{-0.5cm}
\maketitle
\begin{center}
\vspace{-1cm}
Technion - Israel Institute of Technology
\vspace{0.3cm}
\end{center}

\input{sections/abstract}

\input{sections/introduction}

\input{sections/related}

\input{sections/method}

\input{sections/results}

\input{sections/discussion}

\bibliographystyle{plainnat}
\bibliography{references}


\newpage
\section*{NeurIPS Paper Checklist}

\begin{enumerate}

\item {\bf Claims}
    \item[] Question: Do the main claims made in the abstract and introduction accurately reflect the paper's contributions and scope?
    \item[] Answer: \answerYes{} 
    \item[] Justification: All claims made in the abstract and introduction are met throughout the paper. Although some of the goals of the paper are aspirational, we clearly discuss this in Section \ref{sec:discussion}.

\item {\bf Limitations}
    \item[] Question: Does the paper discuss the limitations of the work performed by the authors?
    \item[] Answer: \answerYes{} 
    \item[] Justification: The limitations of our work are discussed in detail in Section \ref{sec:discussion}.

\item {\bf Theory assumptions and proofs}
    \item[] Question: For each theoretical result, does the paper provide the full set of assumptions and a complete (and correct) proof?
    \item[] Answer: \answerNA{} 
    \item[] Justification: 

    \item {\bf Experimental result reproducibility}
    \item[] Question: Does the paper fully disclose all the information needed to reproduce the main experimental results of the paper to the extent that it affects the main claims and/or conclusions of the paper (regardless of whether the code and data are provided or not)?
    \item[] Answer: \answerYes{} 
    \item[] Justification: We provide detailed descriptions of our learning algorithms, control algorithms, network architectures, and hyperparameter choices in the supplementary material. We provide detailed explanations on how we generated our data and fabricated our models, including photos illustrating step-by-step instructions. We will make our date and checkpoints publicly available. We include source code for our simulated experiments in the supplementary material.

\item {\bf Open access to data and code}
    \item[] Question: Does the paper provide open access to the data and code, with sufficient instructions to faithfully reproduce the main experimental results, as described in supplemental material?
    \item[] Answer: \answerYes{} 
    \item[] Justification: The code is publicly available. All of the data, including simulated data, MRI images, and palpation data, have been made open-source.

\item {\bf Experimental setting/details}
    \item[] Question: Does the paper specify all the training and test details (e.g., data splits, hyperparameters, how they were chosen, type of optimizer, etc.) necessary to understand the results?
    \item[] Answer: \answerYes{} 
    \item[] Justification: Most of the details regarding the experiments are provided in the main text while other, more technical details, are presented in \cref{sec:selfsup_details,sec:downstream_tasks_details}.

\item {\bf Experiment statistical significance}
    \item[] Question: Does the paper report error bars suitably and correctly defined or other appropriate information about the statistical significance of the experiments?
    \item[] Answer: \answerYes{} 
    \item[] Justification: All relevant experiments contain error bars, which were calculated by running the experiments with multiple random seeds. 

\item {\bf Experiments compute resources}
    \item[] Question: For each experiment, does the paper provide sufficient information on the computer resources (type of compute workers, memory, time of execution) needed to reproduce the experiments?
    \item[] Answer: \answerYes{} 
    \item[] Justification: The computer resources that were used in the experiments throughout the paper are detailed in Section \ref{sec:compute}.
    
\item {\bf Code of ethics}
    \item[] Question: Does the research conducted in the paper conform, in every respect, with the NeurIPS Code of Ethics \url{https://neurips.cc/public/EthicsGuidelines}?
    \item[] Answer: \answerYes{} 
    \item[] Justification: The paper conforms to the NeurIPS Code of Ethics.

\item {\bf Broader impacts}
    \item[] Question: Does the paper discuss both potential positive societal impacts and negative societal impacts of the work performed?
    \item[] Answer: \answerYes{} 
    \item[] Justification: The paper is a foundational research -- proof of concept for a touch processing framework that can potentially be used in medicine, and we discussed potential positive societal impacts in Section \ref{sec:discussion}. We do not foresee any meaningful negative societal impacts of the work worth discussing.
    
\item {\bf Safeguards}
    \item[] Question: Does the paper describe safeguards that have been put in place for responsible release of data or models that have a high risk for misuse (e.g., pretrained language models, image generators, or scraped datasets)?
    \item[] Answer: \answerNA{} 
    \item[] Justification: \answerNA{}

\item {\bf Licenses for existing assets}
    \item[] Question: Are the creators or original owners of assets (e.g., code, data, models), used in the paper, properly credited and are the license and terms of use explicitly mentioned and properly respected?
    \item[] Answer: \answerYes{} 
    \item[] Justification: we used the panda-py open source package from \cite{elsner2023taming} and credited appropriately in the Results section in the text.

\item {\bf New assets}
    \item[] Question: Are new assets introduced in the paper well documented and is the documentation provided alongside the assets?
    \item[] Answer:  \answerYes{}
    \item[] Justification: The code and data have been released with proper instructions.

\item {\bf Crowdsourcing and research with human subjects}
    \item[] Question: For crowdsourcing experiments and research with human subjects, does the paper include the full text of instructions given to participants and screenshots, if applicable, as well as details about compensation (if any)? 
    \item[] Answer: \answerYes{} 
    \item[] Justification: The detailed procedure of our human-study, including given instructions and financial compensation, can be seen in Section \ref{app:human_study} 

\item {\bf Institutional review board (IRB) approvals or equivalent for research with human subjects}
    \item[] Question: Does the paper describe potential risks incurred by study participants, whether such risks were disclosed to the subjects, and whether Institutional Review Board (IRB) approvals (or an equivalent approval/review based on the requirements of your country or institution) were obtained?
    \item[] Answer: \answerYes{} 
    \item[] Justification: There are no potential risks incurred by study participants, and an IRB has approved it (\cref{app:human_study}).

\item {\bf Declaration of LLM usage}
    \item[] Question: Does the paper describe the usage of LLMs if it is an important, original, or non-standard component of the core methods in this research? Note that if the LLM is used only for writing, editing, or formatting purposes and does not impact the core methodology, scientific rigorousness, or originality of the research, declaration is not required.
    \item[] Answer: \answerNA{} 
    \item[] Justification: 

\end{enumerate}

\newpage

\appendix

\newpage

\input{sections/data_collection}

\input{sections/appendix}

\end{document}

%% file: defs.tex
\newcommand{\spose}{x}
\newcommand{\ttime}{t}
\newcommand{\softbody}{M}
\newcommand{\bodyobs}{I}
\newcommand{\force}{f}
\newcommand{\forcedim}{k}
\newcommand{\trajlength}{T}
\newcommand{\nummodels}{N}
\newcommand{\nummodelobs}{N_\bodyobs}
\newcommand{\data}{\mathcal{D}}
\newcommand{\repr}{z}
\newcommand{\reprdim}{d_z}
\newcommand{\subsample}{K}

%% file: sections/abstract.tex
\begin{abstract}
  Palpation, the use of touch in medical examination, is almost exclusively performed by humans. We investigate a proof of concept for an artificial palpation method based on self-supervised learning. Our key idea is that an encoder-decoder framework can learn a \textit{representation} from a sequence of tactile measurements that contains all the relevant information about the palpated object. We conjecture that such a representation can be used for downstream tasks such as tactile imaging and change detection. With enough training data, it should capture intricate patterns in the tactile measurements that go beyond a simple map of forces -- the current state of the art. To validate our approach, we both develop a simulation environment and collect a real-world dataset of soft objects and corresponding ground truth images obtained by magnetic resonance imaging (MRI). We collect palpation sequences using a robot equipped with a tactile sensor, and train a model that predicts sensory readings at different positions on the object. We investigate the representation learned in this process, and demonstrate its use in imaging and change detection.
\end{abstract}

%% file: sections/introduction.tex
\section{Introduction}
\label{sec:intro}

Palpation, the use of touch in medical examination, is a centuries-old practice that is still important today. In woman's breast cancer -- a case that motivates our work, a large fraction of cases (over 40\%  in the US, per \citealt{roth2011self}) are discovered by palpation, either via self-examination or by a physician, although screening tests based on mammography (X-ray) and magnetic resonance imaging (MRI) are designed to detect tumors that are not yet palpable. The recommendations of the \citet{ACS2023} for women to be familiar with how their breasts normally feel and report any changes promptly indicate that tactile information is still relevant for early detection.

Artificial palpation has the potential to better exploit tactile information by detecting patient-specific temporal changes that physicians typically cannot keep track of, and by improving palpation precision beyond what untrained patients can achieve. Tactile imaging methods \citep{sarvazyan2012tactile} typically involve pressing a force sensor array against soft tissue to generate a force map, which can then be analyzed either visually or via computer vision algorithms. In contrast, elastography~\citep{sarvazyan2011overview} infers the elastic properties and stiffness of soft tissue by measuring variations in ultrasound (US) or magnetic resonance (MR) signals in response to applied forces. Additionally, several studies have investigated the use of tactile sensors for classifying tissue based on stiffness~\citep{jia2013lump,nichols2015methods,di2024using}.

Aimed at improving tactile imaging and detection accuracy, we propose to go beyond direct stiffness estimation and view palpation as an \textit{inference process} of mechanical structures in a soft body, given a sequence of partial, noisy, tactile force measurements. Indeed, a physician performing a breast examination tries to infer the existence of lumps, cysts, and other anatomical structures from touch. The medical literature that characterizes, for example, benign masses as `smooth, soft to firm, and mobile, with well-defined margins'~\citep{klein2005evaluation}, hints that human palpation relies on more involved characterization of how different structures give in to finger \textit{motions} rather than stiffness alone.

While palpation can be taught and perfected, human touch is a \textit{general} skill, learned from years of physical interactions and manipulations of objects. We argue that a data-driven approach to artificial palpation can learn, from touching many different objects, how to interpret a sequence of tactile measurements into corresponding mechanical structures, potentially leading to more accurate imaging than currently available.

Motivated by self-supervised learning results in computer vision and natural language processing~\citep{he2022masked,devlin2019bert}, we propose to learn a general, artificial, palpation \textit{representation}, by predicting future tactile force measurements given a sequence of past measurements. If the representation is useful for predicting future forces, it must contain relevant information about the object being palpated, and may therefore be useful for tactile imaging, change detection, and classification of suspicious findings.

\begin{figure}[ht]
    \centering
    \includegraphics[width=\linewidth]{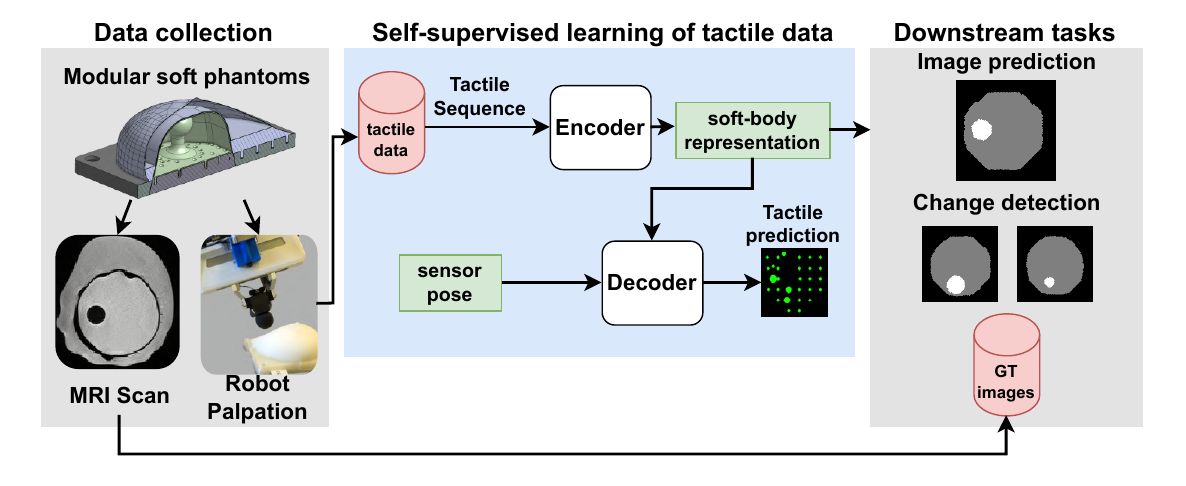}
    \caption{Proof of concept system for learning artificial breast palpation. Left: we fabricate soft objects and palpate them using a tactile sensor mounted on a robot arm. We also obtain MRI scans of the objects as ground truth object models. Middle: we train an encoder-decoder neural network to predict the tactile measurements at given positions from a sequence of previous measurements. Right: we use the learned representation to train a model for tactile imaging, and perform change detection based on predicted images. In principle, by replacing the phantoms with human subjects, our system can be used for clinical studies.}
    \label{fig:overview}
\end{figure}

We present a proof of concept system that includes object fabrication techniques, data collection routines, model training, and an evaluation protocol of tactile imaging and change detection for soft bodies. Our fabricated objects are novel soft breast phantoms with a modular component that can include lumps with various sizes and shapes, resulting in over $1150$ possible
object configurations. We collect data using a robotic manipulator with a tactile sensor tip programmed to palpate the phantom, and train a neural representation by learning to minimize tactile force prediction error. To evaluate the utility of the representation for tactile imaging, we scan the phantoms in an MRI to generate ground truth object models, and train an additional neural network to predict this model from the palpation representation. To evaluate change detection, we use the predicted models to evaluate change in the size of the lump, and compare with human evaluations. We find that our learned representation contains relevant information about the position and shape of the lump, and yields tactile images that are arguably easier to interpret than a map of forces, which can be used for change detection at a level comparable to humans.

While clearly only a prototype, we argue that these results provide a promising direction for an artificial palpation system that uses human MRI scans and tactile data to \textit{learn} tactile imaging.

%% file: sections/related.tex
\section{Related Work}
\textbf{Learning tactile representations:} work in this area focused almost exclusively on rigid bodies, and applications to object manipulation and shape reconstruction. 
\citet{guzey2023dexterity} considered instantaneous tactile measurements, and applied BYOL~\citep{grill2020bootstrap} to multiple taxel readings arranged as an image. \citet{higuera2024sparsh} considered vision-based tactile sensors~\citep{lambeta2020digit}, proposed several self-supervised learning representations, and a benchmark for evaluating them in various robotic manipulating tasks. More recently, \citet{feng2025anytouch} collected matched tactile data for four different sensors, and proposed an encoder-decoder framework for images and short ($3$-frame) videos from vision-based tactile sensors.
\citet{qi2023general} represented a sequence of tactile, proprioceptive, and visual observations of a rigid body, by learning to reconstruct simulated 3D point-clouds of the object. The representation was further used for learning in-hand object rotation and shape reconstruction. Focusing on rigid-object shape reconstruction and pose estimation, \cite{zhao2023fingerslam} estimate pose and shape with neural networks, from data of manually pressing a rigid object with a visuo-tactile sensor, while \citet{suresh2024neuralfeels} used neural fields to learn the pose and 3D shape of an object rotated by a robotic hand. Neither of these methods are applicable to estimating the properties of soft objects.

\textbf{Robotic Palpation and Cancer Diagnosis} \citet{khanna2024robotics} provide a recent survey in the context of breast cancer diagnosis; we highlight several studies most relevant to our work. 
\citet{jenkinson2023robotic} develop a radial robotic mechanism for breast palpation, while \citet{syrymova2025breast} investigate a purpose-built tactile glove.
\citet{scimeca2022action} palpate a soft silicone object with hard inclusions of 3 different sizes, and classify the size of the inclusion by projecting instantaneous measurements to their first principal component, and fitting a Gaussian density to the projected measurements over time per inclusion size. Our work is similar in spirit but significantly larger in scale, both in the data and the learning models. \citet{di2024using} used data from a vision-based tactile sensor to classify both prostate phantoms and real prostate tissue ex-vivo according their hardness, by fine-tuning a video masked auto-encoder applied to sequences of images.

Several studies explored how to define the robot's \textit{motion} during palpation. \citet{scimeca2022action} investigated various motions that revolve around a point in space. \citet{sanni2022deep} used deep movement primitives for robotic breast palpation, while \citet{zhao2025remote} explored a shared autonomy scheme between a human tele-operator and computerized control. In this work we focus on simple linear motions and leave the investigation of more complex movements to future work.

In terms of phantom fabrication, most of the literature concerns fabrications suitable for MRI and ultrasound imaging~\citep[e.g.,][]{ustbas2018silicone,keenan2016design}. We are not aware of standard methods for fabricating phantoms for breast palpation.

%% file: sections/method.tex
\vspace{-0.5em}
\section{Method}\label{sec:method}
\vspace{-0.5em}
We propose a data-driven approach for soft-body tactile palpation. We first formulate our inference problem. We then use self-supervised learning approaches to learn a representation of the palpated object
(\cref{sec:method_rep}). Finally, we use this representation to learn downstream tasks with limited amount of supervised data (\cref{sec:method_imaging,sec:method_change}).

\vspace{-1em}
\subsection{Formulation} \label{sec:method_formulation}
\begin{wrapfigure}{r}{0.35\textwidth}
\vspace{-1.5em}
\scalebox{0.7}{
\begin{tikzpicture}[
node distance=1.3cm and 1.3cm,
every node/.style={draw, circle, minimum size=1cm, font=\large, line width=1pt},
>=Stealth
]

\node (M) at (0,0) {$\softbody$};
\node (Xt) [right=of M] {$\spose_t$};
\node (Xt1) [right=of Xt] {$\spose_{t+1}$};
\node (ft) [below=of Xt] {$\force_t$};
\node (ft1) [below=of Xt1] {$\force_{t+1}$};
\node (I) [below=of M] {$\bodyobs$};

\draw[->, line width=0.6mm] (Xt) -- (Xt1);
\draw[->, line width=0.6mm] (Xt) -- (ft);
\draw[->, line width=0.6mm] (Xt1) -- (ft1);
\draw[->, line width=0.6mm] (M) -- (ft);
\draw[->, line width=0.6mm] (M) -- (ft1);
\draw[->, line width=0.6mm] (M) -- (I);
\draw[dashed,->, line width=0.6mm] (Xt1) -- ++(1.3,0);
\draw[dashed,->, line width=0.6mm] (ft1) -- ++(1.3,0);

\end{tikzpicture}
}
\vspace{-1em}
\end{wrapfigure}
We model the artificial palpation problem as follows.
A tactile sensor is a rigid body that is controlled to be at time $\ttime \in 0,\dots,\trajlength$, in pose $\spose_\ttime \in \mathbb{R}^{6}$.\footnote{For simplicity we consider a single sensor; our formulation trivially extends to multiple sensors.} 
The sensor interacts with a soft body $\softbody$, resulting in a $\forcedim$-dimensional force reading $\force_\ttime \in \mathbb{R}^{\forcedim}$. We consider $\softbody$ to represent all the structural and mechanical properties of the body, which determine the force on the sensor, and note that in all practical situations $\softbody$ is unknown. Subsequently, the sensor is moved to the next pose $\spose_{\ttime+1}$ by the controller, and the next reading is obtained. In addition, we assume access to some observation of the soft body, denoted $\bodyobs$, for example, an MRI scan. In this work we do not consider \textit{how} to control the sensor, and assume that some palpation motion controller is available. Also, we assume that the force sensor is noisy, but we do not model the noise in any explicit form.

In the \texttt{tactile imaging} problem, our goal is to use the poses and force readings $\spose_0,\force_0,\dots, \spose_\trajlength,\force_\trajlength$ to predict the observation $\bodyobs$. In the \texttt{change detection} problem, we are given readings from two bodies $\softbody, \softbody'$, which may or may not be different, and our goal is to determine if $\softbody = \softbody'$.

\subsection{Representation Learning} \label{sec:method_rep}

To address the inference problems described above, we propose a learning based approach. We first explain the data structure, and then our learning algorithm.

We collect data from $\nummodels$ bodies $\softbody_1,\dots,\softbody_{\nummodels}$. For each body $\softbody_i$, we collect tactile data $\data_i \equiv \left\{\spose_\ttime^i, \force_\ttime^i\right\}_{\ttime=1}^{\trajlength}$, by a palpation controller as described above. In addition, for the first $\nummodelobs$ bodies we have a corresponding observation $\bodyobs_i$, $i \leq \nummodelobs$. 

\begin{figure}
    \centering
    \includegraphics[width=0.95\linewidth]{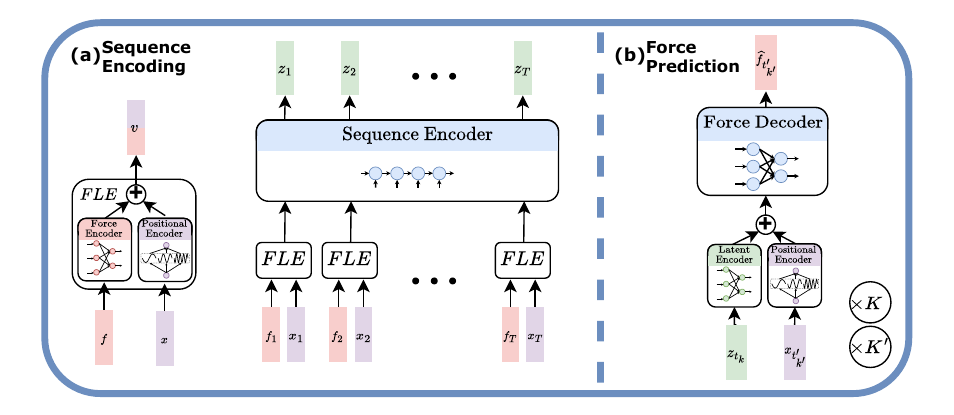}
    \caption{\textbf{Representation Learning:} (a) A sequence of tactile measurements and poses is encoded by first encoding every measurement+pose by a force-location encoder (FLE), and then encoding the sequence by a GRU. 
    (b) The decoder predicts a tactile measurement at time $\ttime'_{k'}$ from the representation at time $\ttime_k$ and the pose at time $\ttime'_{k'}$.}
    \label{fig:method_rep}
    \vspace{-1em}
\end{figure}

Our working hypothesis is that collecting large amounts of observations (e.g., MRI scans) is more difficult than collecting palpation data, that is, $\nummodelobs < \nummodels$. Following prior works~\citep{devlin2019bert,he2022masked}, we design our approach to first use self-supervised learning techniques to pretrain a representation from the large amount of palpation data, and later use the pretrained representation to effectively learn downstream tasks such as \texttt{tactile imaging}.

We use an encoder-decoder architecture to learn a representation in a self-supervised manner. The key idea is that the encoder is trained to map a sequence of $t$ measurements $\left\{\spose_\ttime^i, \force_\ttime^i\right\}_{\ttime=1}^{t}$ from a body $\softbody_i$ to a $\reprdim$-dimensional vector representation $\repr_t \in \mathbb{R}^{\reprdim}$ that contains the information in the measurements about $\softbody_i$. Thus, at the last time step $\trajlength$, the representation $\repr_{\trajlength}$ contains information about the complete trajectory,
and can be used for downstream tasks. We next detail our architecture and training objective.

\textbf{Encoder}

Our encoder maps between a sequence of measurements 
to a sequence of latent representations. We first use a force-location encoder (FLE) to encode each step separately:
\begin{small}
$
FLE(\force_t, \spose_t) = MLP(\force_t) + PE(\spose_t), \qquad \forall t \in \left[1, \trajlength\right].
$
\end{small}

The forces are encoded via a two-layer MLP, and the locations are encoded with a sinusoidal positional encoding (PE). 
Afterwards, we use a sequence encoder to produce a sequence of embeddings of the same length as the input sequence $\left\{\repr_t\right\}_{t=1}^{\trajlength}$.
We choose a gated recurrent unit (GRU) \citep{gru} encoder as it can easily scale to long sequences, and acts as an information bottleneck in the encoding process.\footnote{In simulation experiments that are not reported here, we also investigated other architectures such as transformer-based masked autoencoders~\citep{he2022masked}, but obtained similar results.}

\textbf{Decoder}
We hypothesize that if the representation is informative for predicting force measurements, it should contain relevant information about the body being palpated. Therefore, we structure our force decoder (FD) to predict the force reading at pose $\spose_{t'}$ based on the representation $\repr_t$:
\begin{small}
$
FD\left(\repr_t, \spose_{t'}\right) = MLP(MLP(\repr_t) + PE(\spose_{t'})) \in \mathbb{R}^{\forcedim},\qquad \forall t,{t'} \in \left[1, \trajlength\right],
$
\end{small}
For ${t'} > t$, this corresponds to predicting future sensor readings, while for ${t'} < t$ this means recalling past measurements. We note that a similar idea of predicting both past and future measurements was proposed by~\citet{zintgraf2021varibad} for decision making in partially observed domains.

\paragraph{Training Objective}
We use the mean squared error (MSE) reconstruction loss between the predicted and true forces to train our encoder-decoder model. Since predicting all the forces from all the timestamps has 
an $\mathcal{O}\left(\trajlength^2\right)$ complexity, applying it to long sequences is challenging. Instead, we uniformly subsample reconstruction steps for calculating the loss, as described next for data from a single body $\softbody$: 

\vspace{-1.5em}
\begin{small}
$$
\mathcal{L}_{rec} = \frac{1}{2\subsample \subsample'}  \sum_{k=1}^\subsample 
\sum_{k'=1}^{\subsample'}
\norm{FD\left(\repr_{t_k}, \spose_{t'_{k'}}\right) - \force_{t'_{k'}}}^2,
$$
\end{small}

where $\left\{t_k\right\}_{k=1}^\subsample$ are $\subsample < \trajlength$ uniform samples without replacement from $\left[1, \trajlength\right]$, and for each $k \in \left[1, K\right]$, $\left\{t^{\prime}_{k'}\right\}_{k'=1}^{\subsample'}$ are also $\subsample' < \trajlength$ uniform samples without replacement from $\left[1, \trajlength\right]$. After calculating all representations $\repr_1,\dots, \repr_\trajlength$ during the encoder's forward pass, a decoder forward pass and loss calculation can be done in parallel for all indices $k,k'$ and has a complexity of only $\mathcal{O}\left(\subsample'\cdot \subsample'\right)$. In our implementation we set $\subsample= \subsample'=64$.

\subsection{Tactile Imaging}
\label{sec:method_imaging}
\label{sec:method_change}

We hypothesize that the representations $\repr$ learned as described in Section \ref{sec:method_rep} are useful for predicting more general properties of the body $\softbody$ than forces. In particular, we focus on the \texttt{tactile imaging} problem, and propose to predict the observation $\bodyobs$ from the representation of a complete palpation sequence $\repr_\trajlength$.

We use an image prediction network with an architecture inspired by flow matching \citep{flowmatching}. The network maps the vector representation  $\repr_\trajlength$ and a normally distributed random noise to a $128 \times 128$ image, where each pixel can take one of $3$ values (described later). We train this using standard cross-entropy loss per pixel, based on the ground truth MRI images. We found that adding noise input stabilized training, while the final network was near deterministic (the standard deviation over results from different noise samples is two orders of magnitude smaller than the expectation). In inference, we draw a single noise sample to generate an image. Full technical details are in \cref{sec:downstream_tasks_details}. 
This architecture can easily be modified to handle more general images, or even $3$-dimensional volumetric images. We train the imaging network separately from the pretrained representation. 

For change detection based on the predicted images, we consider two different methods. The first compares between the pixel values of images obtained from palpation trajectories of two bodies directly. This method, which is reported in the supplementary material, works well when the image predictions are relatively accurate, as we obtained in our simulation results. For our real world results, however, we first evaluated the lump size from the predicted images, and detected change based on the predicted lump size (see Section \ref{sec:results}).

%% file: sections/results.tex
\section{Results}\label{sec:results}
We next present our experimental results. We begin with introducing a new simulated environment for palpation, in which we investigate our design choices and quantify the scale of data required to obtain meaningful results. We then detail our real-world data collection and experiments. We provide supplementary visualizations in \href{https://zoharri.github.io/artificial-palpation/}{\texttt{zoharri.github.io/artificial-palpation}}, and our code can be found in the supplementary material.

\subsection{The PalpationSim Simulator} \label{sec:datacollection_sim}

We introduce \texttt{PalpationSim} -- a simulation environment to be used as a mock of the real palpation learning setup. We designed \texttt{PalpationSim} to be both lightweight, quick, and easy to visualize and interpret, yet indicative of our real-world domain. 

\begin{figure}
     \centering
     \begin{subfigure}[b]{0.325\textwidth}
         \centering
         \vspace{0.5em}
         \includegraphics[width=\textwidth]{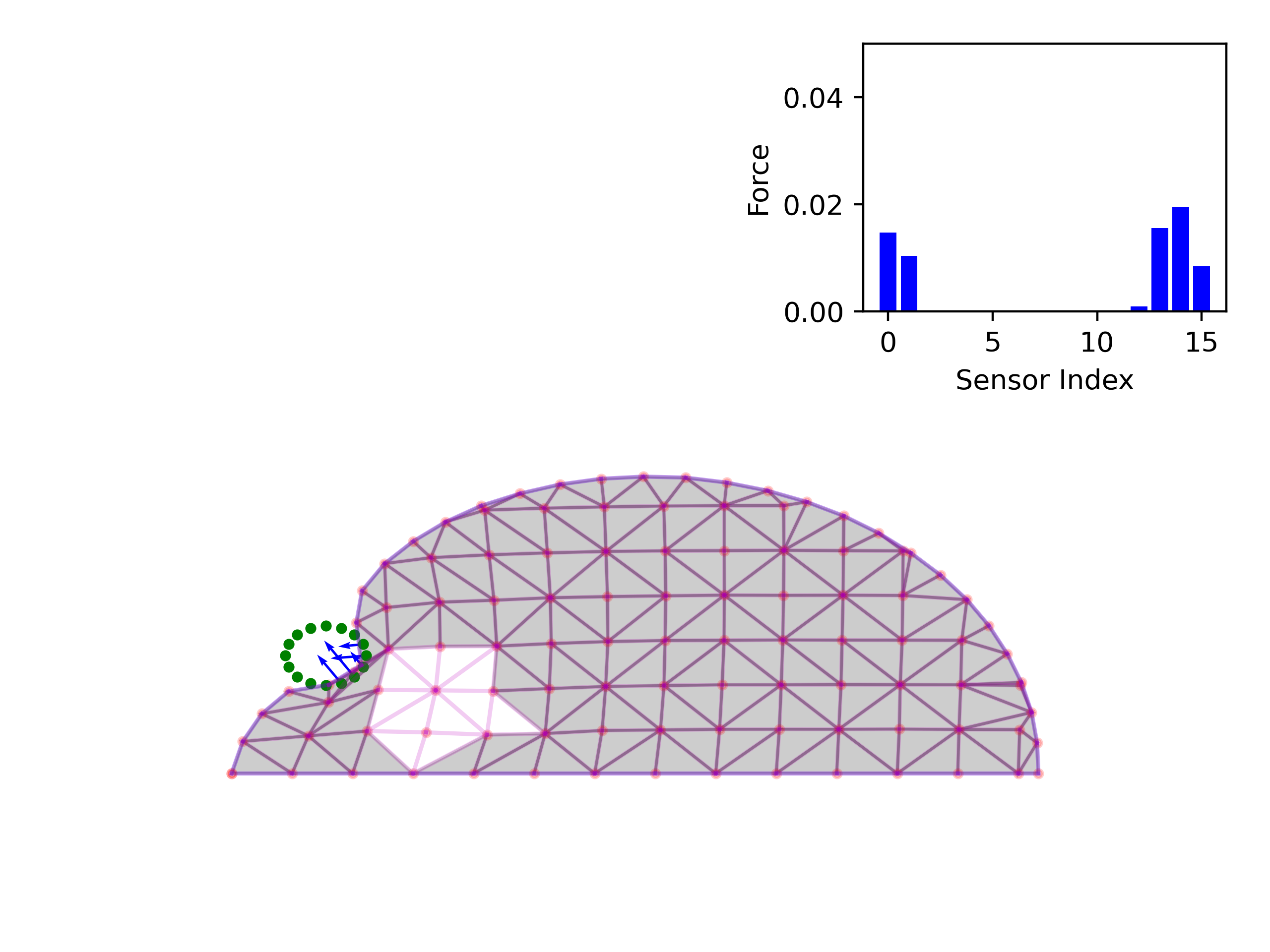}
         \vspace{-3em}
         \caption{}
         \label{fig:sim_fem}
     \end{subfigure}
     \hfill
     \begin{subfigure}[b]{0.325\textwidth}
         \centering
         \includegraphics[width=\textwidth]{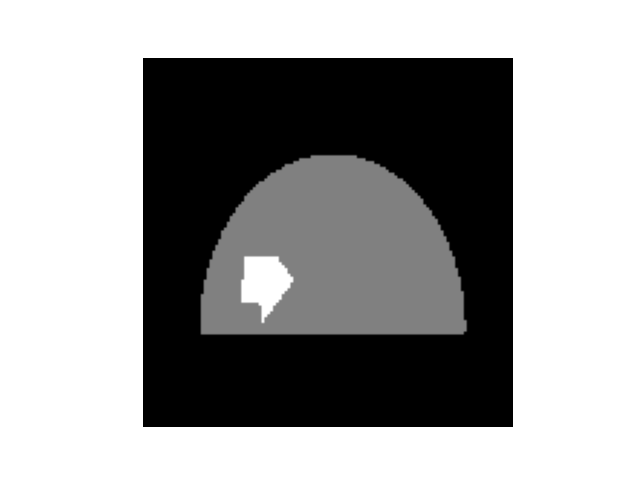}
         \vspace{-2.5em}
         \caption{}
         \label{fig:sim_gt}
     \end{subfigure}
     \hfill
     \begin{subfigure}[b]{0.325\textwidth}
         \centering
         \includegraphics[width=\textwidth]{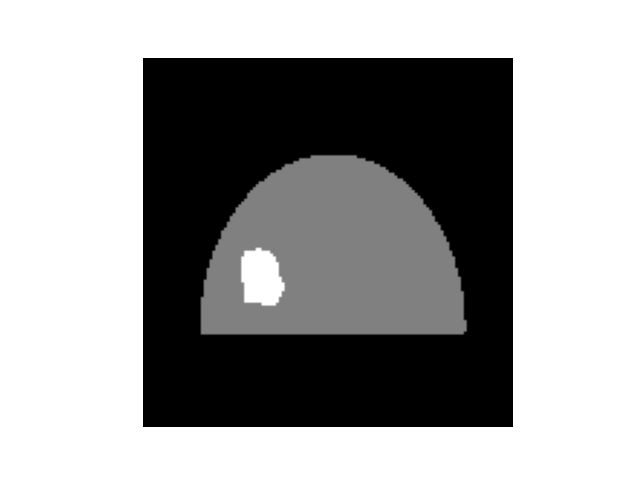}
         \vspace{-2.5em}
         \caption{}
         \label{fig:sim_pred}
     \end{subfigure}
     \\
     \begin{subfigure}[b]{0.325\textwidth}
         \centering
         \includegraphics[width=\textwidth]{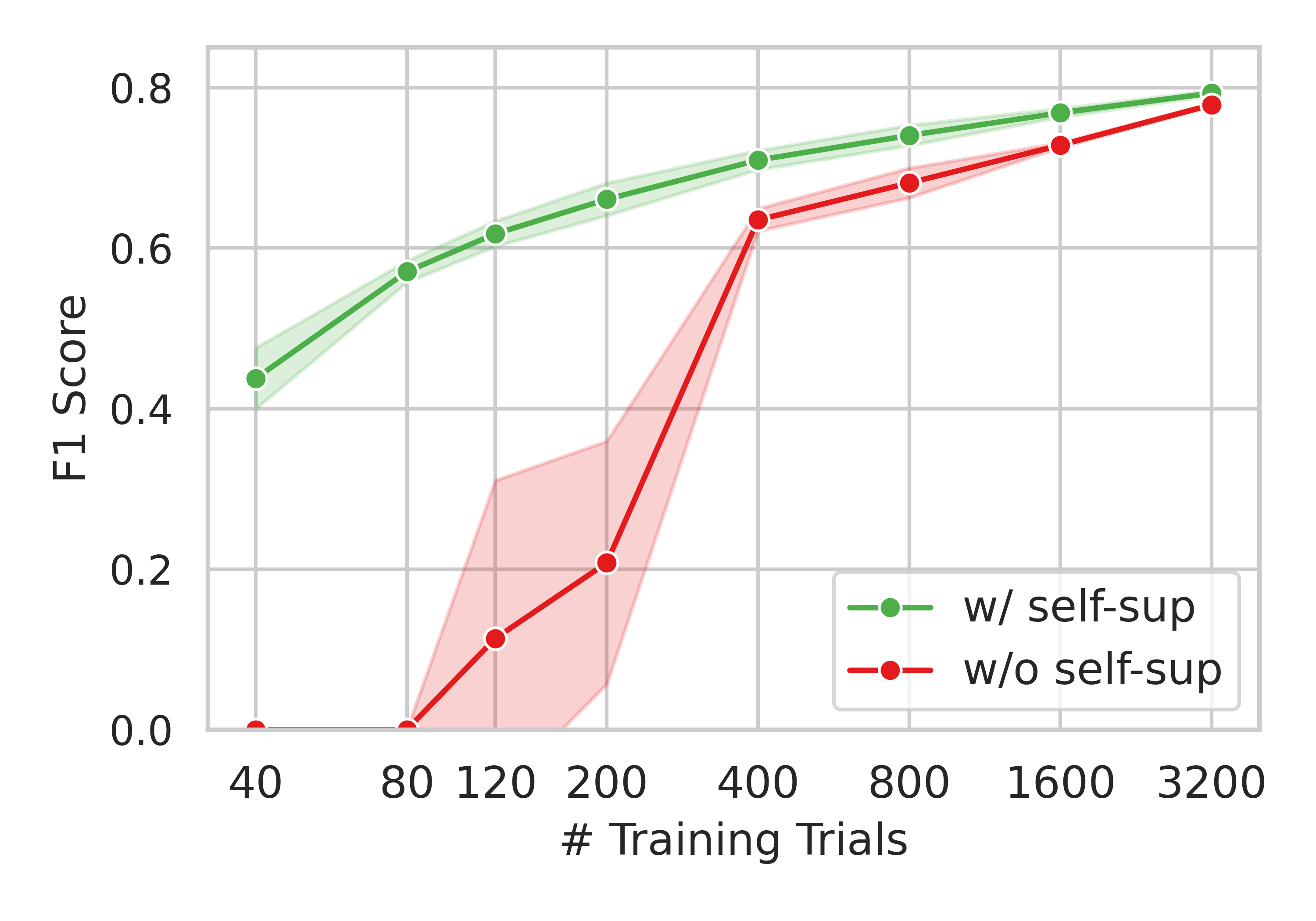}
         \vspace{-2em}
         \caption{}
         \label{fig:sup_unsup}
     \end{subfigure}
     \hfill
     \begin{subfigure}[b]{0.325\textwidth}
         \centering
         \includegraphics[width=\textwidth]{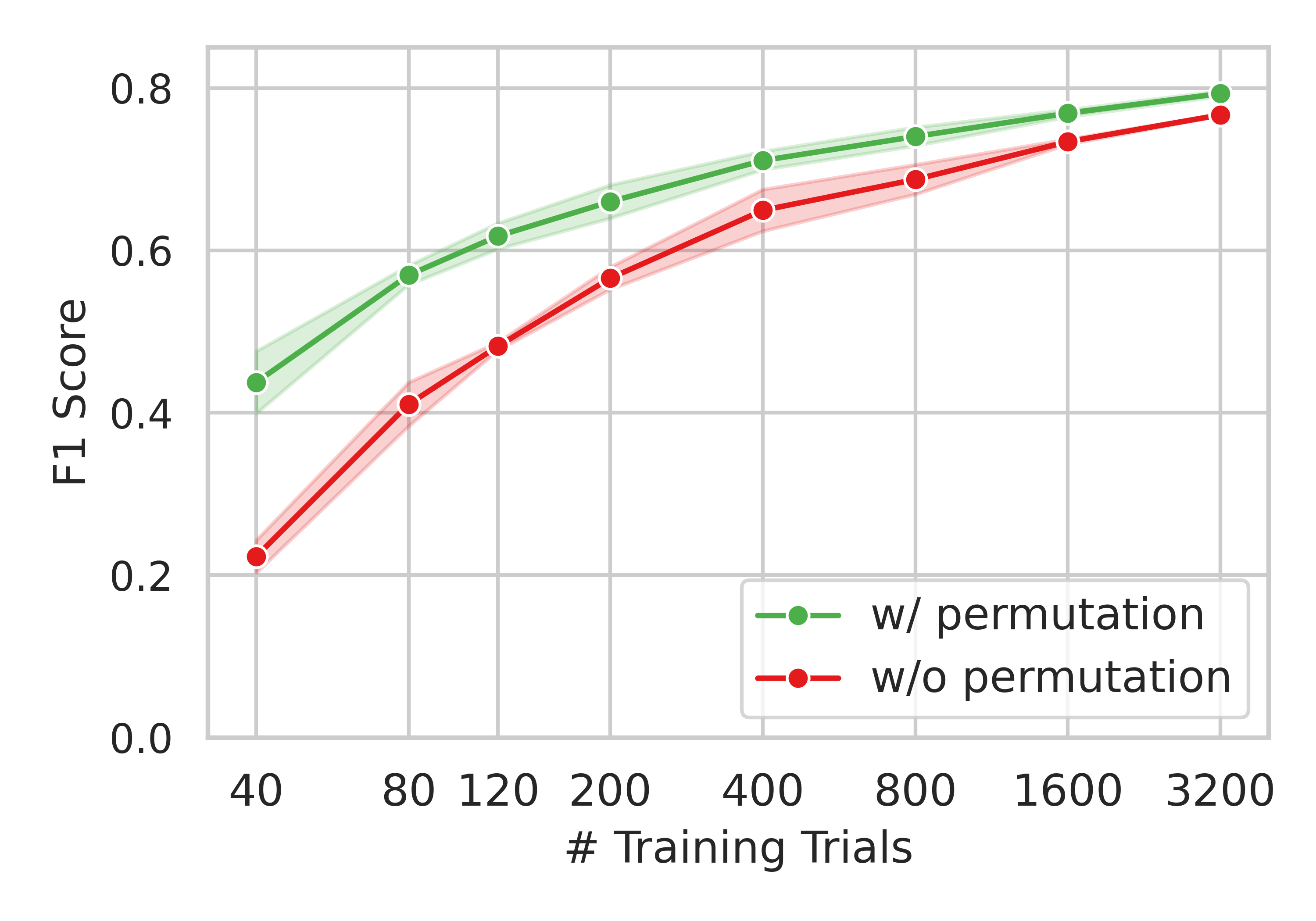}
         \vspace{-2em}
         \caption{}
         \label{fig:sup_permutation_aug}
     \end{subfigure}
     \hfill
      \begin{subfigure}[b]{0.325\textwidth}
         \centering
         \includegraphics[width=\textwidth]{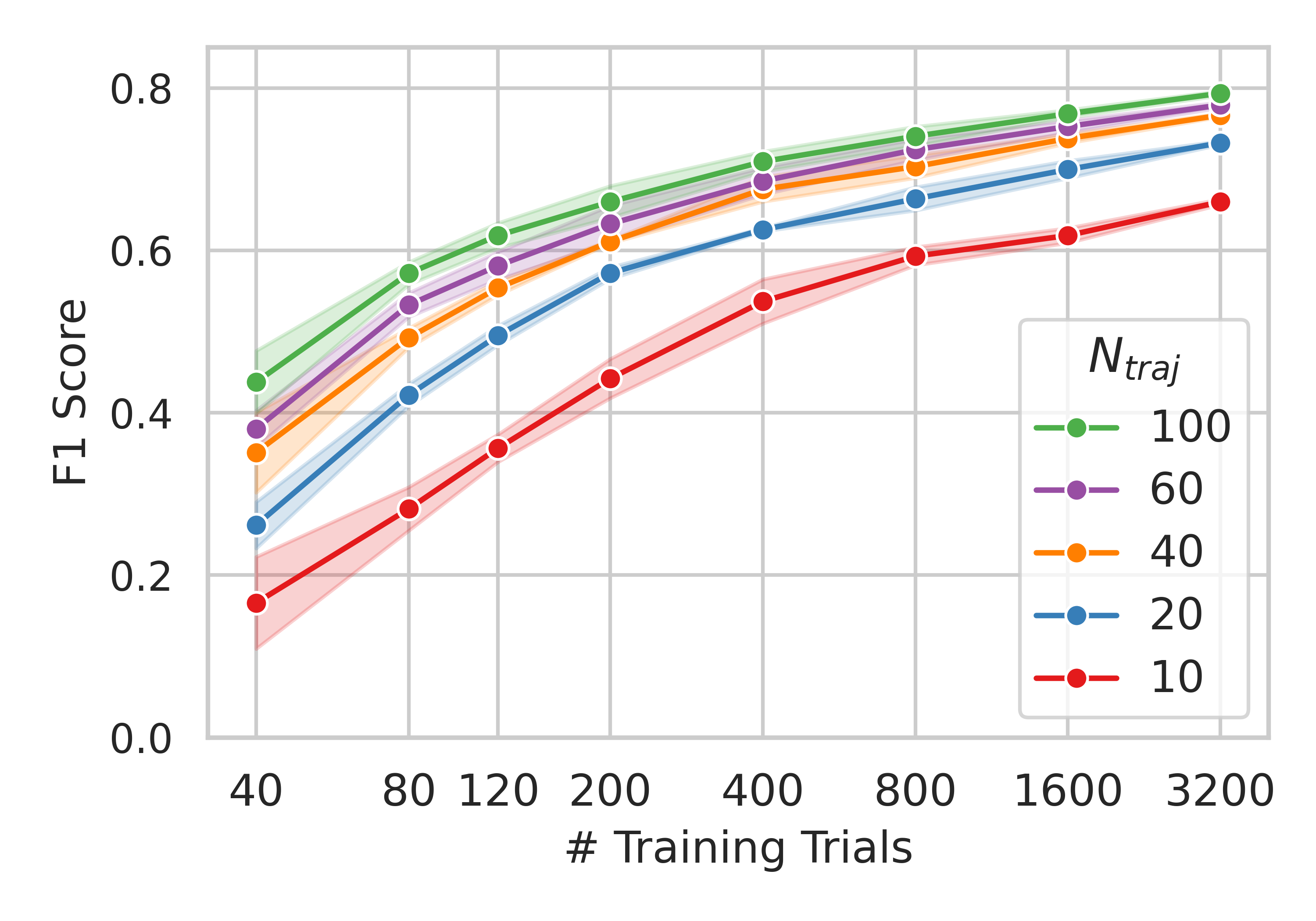}
         \vspace{-2em}
         \caption{}
         \label{fig:num_train_traj}
     \end{subfigure}
        \caption{\texttt{PalpationSim} simulator and simulation results. (a) A $2$-dimensional finite-element model of a round sensor pressing on a soft object with a harder lump inside; insert shows the forces on the sensor. (b+c) A ground truth image of the body (b), and a predicted image (c). (d-f) Image prediction results: (d) with and without self-supervised pretraining, (e) with and without permutation augmentation, (f) with different number of trajectories per trial. See text for details.}
        \label{fig:three graphs}
\end{figure}

Our main component is a 2-dimensional finite element method (FEM) simulation of a soft semi-circular body with the option of having a harder lump element inside, as depicted in Figure \ref{fig:sim_fem}. \footnote{We release \texttt{PalpationSim} as part of our open-source code. While several physical simulators that are popular in the robotics literature can simulate soft objects~\citep{todorov2012mujoco,isaaclab2025}, we did not find any versatile enough for our simulation environment at the time of writing. On the other hand, we desired a much faster and lightweight solution than full blown commercial FEM packages.} The body is composed of triangular elements with linear elasticity, and we model the different hardness of the body and lump by different values of Poisson's ratio and Young's modulus~\citep{lubliner2016introduction}. The bottom vertices of the semi-circle are fixed in place, while the other vertices are free to move. A tactile sensor is modeled using $16$ $2$-dimensional points arranged in a circle, with the center of the circle denoting its pose $\spose$ (here, $\spose$ is a $2$-dimensional position instead of a $6$-dimensional pose). Each point, when inside a triangular element, applies a spring-like repulsion force to the nearest edge of the triangle. Given $\spose$, we find the positions of all the vertices by minimizing the energy of the system using Adam~\citep{kingma2014adam}, and measure the forces on each sensor proportionally to their penetration into the body. Note that the simulation is \textit{quasi-static} -- per sensor position $\spose$, we measure the forces after the vertices have stabilized in a steady state. When we move the sensor, we measure the steady state of the system for each sensor position along the way (we warm-start the energy minimization with the previous solution for faster optimization). This design choice corresponds to a slow motion of the sensor in the real world.

The motions available to the probe $\spose_1,\dots,\spose_{\trajlength}$ are linear trajectories that ``press'' on the soft body at various positions. The body observations $\bodyobs$ are 2-dimensional $128 \times 128$ images where we discretize each pixel value into $3$ classes: background, body, and lump, as depicted in Figures \ref{fig:sim_gt},\ref{fig:sim_pred}.

\paragraph{Data Collection}
We first randomly sample $N_p$ random bodies with different lump locations (if any), sizes, and variations in the Young modulus and Poisson ratio for each finite element. For each body, we collect $N_{trial}=2$ “trials”, where each trial contains a sequence of $N_{traj}$ trajectories pressing on the body from uniformly distributed angles. Between trials, we randomly modify the Poisson's ratio and Young's modulus for each finite element, and, in $10$\% of the models, we increase the size of the lump, for our change detection task. Full details for the data collection appear in Appendix~\ref{app:simulation_data_collection}. 

\paragraph{Simulation Results}
We focus on the following two questions: (1) how important is self-supervised pretraining, and (2) how do results scale with the number of models and the number of trajectories collected from each model. In addition, we report on a simple and useful augmentation.

\cref{fig:sim_gt,fig:sim_pred} demonstrate our image prediction results. In the following, we treat each pixel in the predicted image as a classification problem, and report the $F1$ score for the complete image~\citep{manning2008ir}, which we found to correlate well with visual image quality. We report results for different sizes of the data, per the total number of trials in the dataset (specifically, we train the models on a subsample of trials and corresponding images to report each data point in the figures). In Figure \ref{fig:sup_unsup} we compare the image prediction using a pretrained representation (trained on the full data), as described in \cref{sec:method}, with a supervised learning method that directly predicts the image from a sequence of measurements. Interestingly, pretraining improves results for all sizes of data, and is more dramatic when less data is available, showing the benefit of pretraining with large amounts of only tactile data. In Figure~\ref{fig:num_train_traj}, we show the scaling with respect to number of trajectories from each body used during pretraining; intuitively, performance plateaus when the trajectories sufficiently `cover' the body. Finally, in Figure~\ref{fig:sup_permutation_aug} we show that a simple augmentation of randomly changing the order of trajectories from the same trial during pretraining dramatically improves performance; we used this augmentation in all other reported results.

To conclude, the simulation study allowed us to dimension our real world study, and obtain a well-performing working point for our learning algorithms. We next report our real world results.

\vspace{-0.5em}
\subsection{Real-World Results}
Our goal is a proof-of-concept for artificial palpation. To this end, we designed an experimental procedure that, while artificial and simplistic in nature, addresses some of the realistic challenges in obtaining tactile measurements and ground truth imaging for breast palpation.

\subsubsection{Benchmark Curation}
\begin{figure}
     \centering
     \begin{subfigure}[b]{0.2\textwidth}
         \centering
         \includegraphics[width=\textwidth]{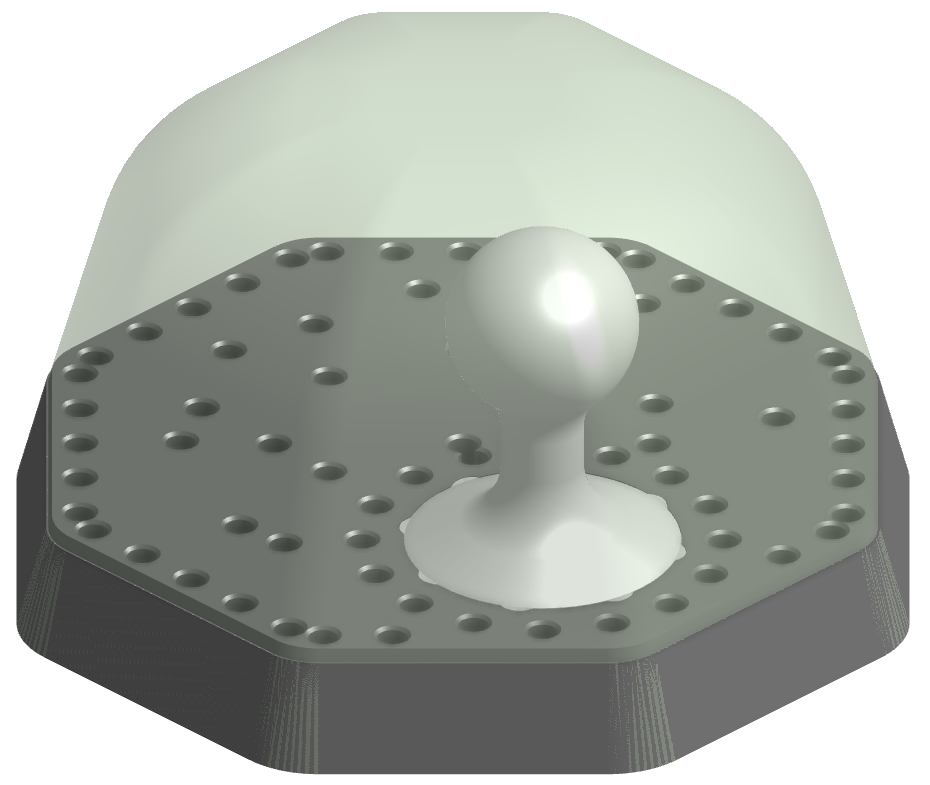}
         \caption{}
         \label{fig:insert_transluscent}
     \end{subfigure}
     \hfill
     \begin{subfigure}[b]{0.3\textwidth}
         \centering
         \includegraphics[width=\textwidth]{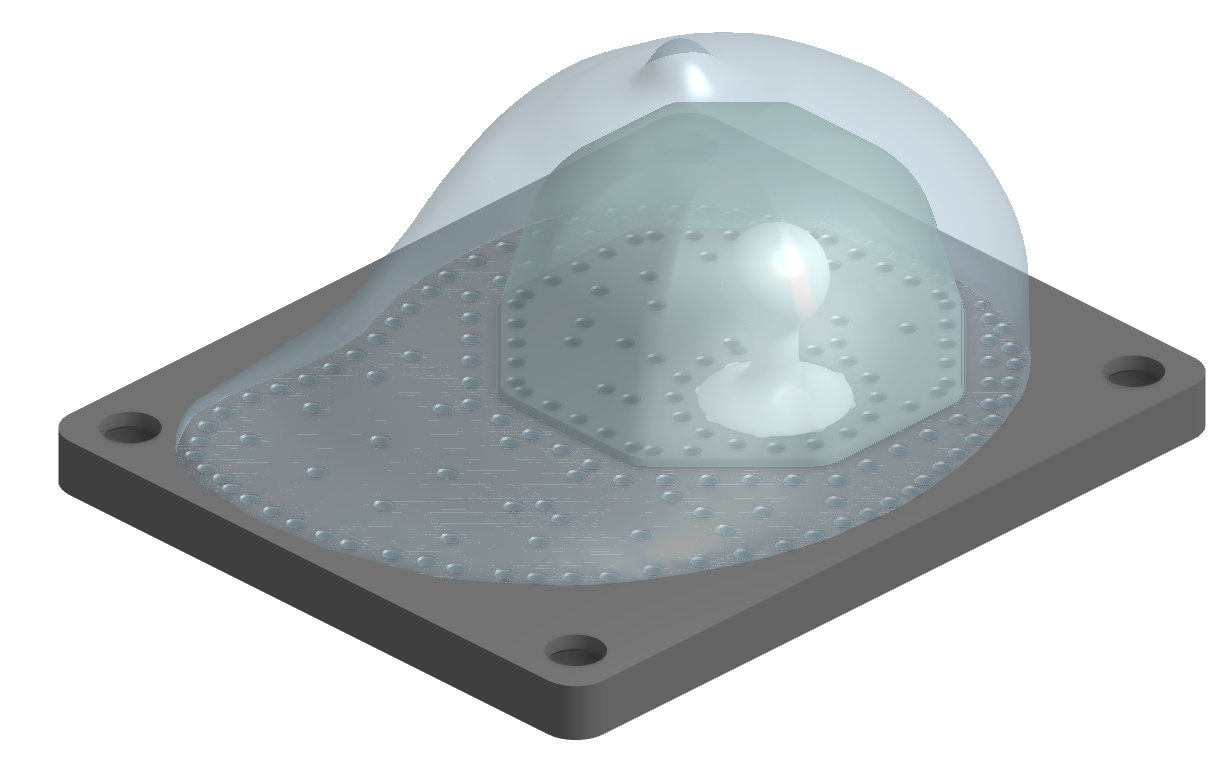}
         \caption{}
         \label{fig:breast_transluscent}
     \end{subfigure}
     \hfill
      \begin{subfigure}[b]{0.38\textwidth}
         \centering
         \includegraphics[width=\textwidth]{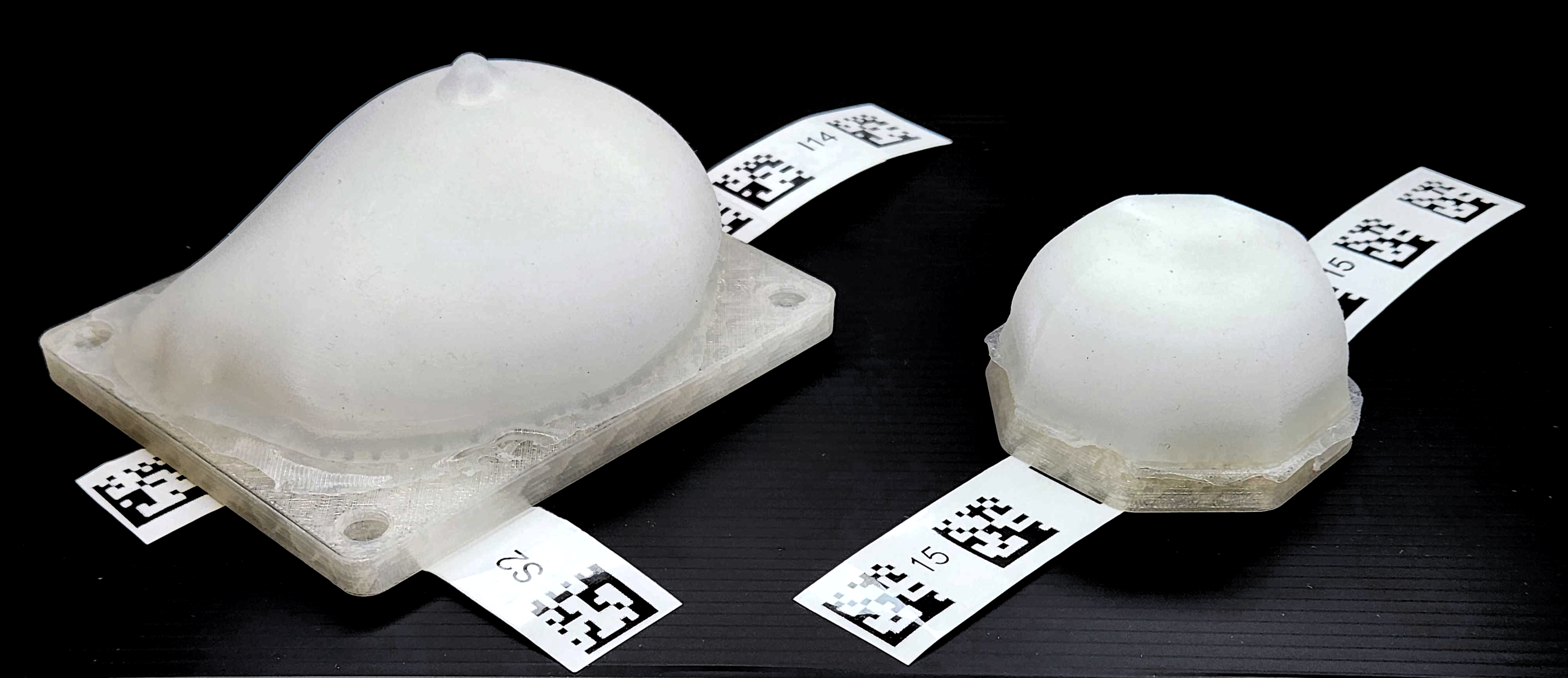}
         \caption{}
         \label{fig:phantom_black}
     \end{subfigure}
        \caption{Modular Breast Phantom Design. (a) The \textit{insert} has an octagonal 3D-printed base, a soft-silicone skin, and is filled with polyvinyl acetate hydrogel. A soft silicone ``lump'' is embedded within and attached to the base. (b) The \textit{shell} has two layers of soft silicone skin, with hydrogel in between, where the bottom layer is attached to a 3D-printed base with an octagonal hole. The insert can be positioned in $8$ possible orientations inside the shell. (c) An assembled shell+insert and a standalone insert. The bar-code labels allow to automatically record the component types and orientations using an overhead camera.}
        \label{fig:phantom_main}
        \vspace{-1em}
\end{figure}

\textbf{Modular Breast Phantoms:} At the core of our experiments is a novel breast phantom design (see \cref{fig:phantom_main}) that is composed of two modular parts, an outer \textit{shell} and an \textit{insert}, which can be assembled together in multiple orientations to collect data from a variety of bodies. Each part is composed of a soft silicone skin, and a polyvinyl acetate hydrogel filling, mimicking the breast fat tissue. We add a lump of certain shape inside the insert made also of soft silicone. While our phantoms are clearly not anatomically accurate depictions of human breasts, we posit that they reflect a realistic range of tactile sensations present in breast palpation, as we verified with several physicians and breast oncologists. We provide a detailed description of the phantom and instructions for reproducing our design in Appendix~\ref{sec:datacollection}. We fabricated $6$
shells with different hydrogel characteristics and inserts with $8$ different lumps positions and shapes, each in $3$ different sizes, ranging from $8$[mm] to $14$[mm] diameter, which can be placed inside the shell in $8$ different orientations, resulting in over $1150$ possible different phantom combinations that we can collect data with.

\textbf{Automatic Palpation Data Collection:} 
We use a single XELA uSkin tactile sensor~\citep{xelauskin}, 
which has $30$ $3$-dimensional force sensors, and records data at $85$ Hz.\footnote{We opted for the uSkin based on a preliminary comparison with a Digit sensor~\citep{lambeta2020digit}, reported in detail in \cref{sec:digit_vs_xela}. We found the vision-based Digit to perform poorly inside the soft material, which we conjecture is due to the low spatial frequency of relevant forces, which the Digit is not optimized for.}

To collect data in a consistent and repeatable manner, we use a Franka Emika Panda robotic manipulator with the tactile sensor attached to its end effector, as depicted in \cref{fig:robotic_setup}. We control the arm using a hybrid force-motion controller~\citep{lynch2017modern} that applies a downward motion of the tip and a vertical force of $3.8$N at a fixed orientation, implemented as a modification of the default force controller in panda-py~\citep{elsner2023taming}. Such a downward motion at a particular $x-y$ position lasting for $5$ seconds is termed a `poke'. We execute pokes on a preset matrix of $110$ positions, and for each poke we record both the sensor readings, the robot end effector poses, and their corresponding timestamps. In total, collecting data from a single phantom takes $20$ minutes of robot time.
Collecting data from $\sim 550$ phantom combinations resulted in a total of $\sim 60 K$ pokes, and $\sim 30 M$ instantaneous sensor readings. To mitigate time-varying bias in the Xela sensor, each measurement sequence was normalized relative to its first sensor reading.

\paragraph{Ground Truth Phantom MR Images:} We generate ground truth labels for tactile imaging by scanning our phantoms in an MRI system. The motivation for this is threefold: (1) while we have 3D CAD designs for the phantom molds, our manufacturing process of the silicone skin has inaccuracies due to its manual nature; (2) the MRI scans, while not realistic due to the simple structure of our phantoms, are still susceptible to real-world challenges of noise and variations in positions and shape due to the softness of the phantom, making for a more challenging prediction problem; and (3) for a future human study, collecting ground truth data using MRI is a viable approach.

We scan our inserts\footnote{We scan inserts instead of the complete phantoms for time and cost reasons -- we scan several inserts at once, and rotate the insert images digitally, obtaining scans for almost $200$ different bodies. While removing the shell makes for a slightly easier prediction task, we found it challenging enough to obtain meaningful insights} using a $3$T MRI (Siemens Prisma) system with a 64-channel coil. The acquisition protocol involved a volumetric T2-weighted SPACE (fast spin echo \citep{bernstein2004handbook}) acquisition. The parameters were a turbo factor of 270,  resolution of $0.8 \times 0.8 \times 0.9$, field of view was $220 \times 200 \times 86.4 mm$,  TR of $3200ms$, and TE of $412ms$. The scans were accelerated by 4x using a standard undersampling technique. The images were reconstructed on the scanner using the GRAPPA algorithm \citep{griswold2002generalized}. They were then imported to an external servers for training the networks. Since in our inserts lumps have similar height, for simplicity we predict a horizontal image slice instead of a complete 3D model (full details are in \cref{sec:mri_preprocessing}).

\subsubsection{Results}

\begin{figure}
    \centering
    \includegraphics[width=0.9\linewidth]{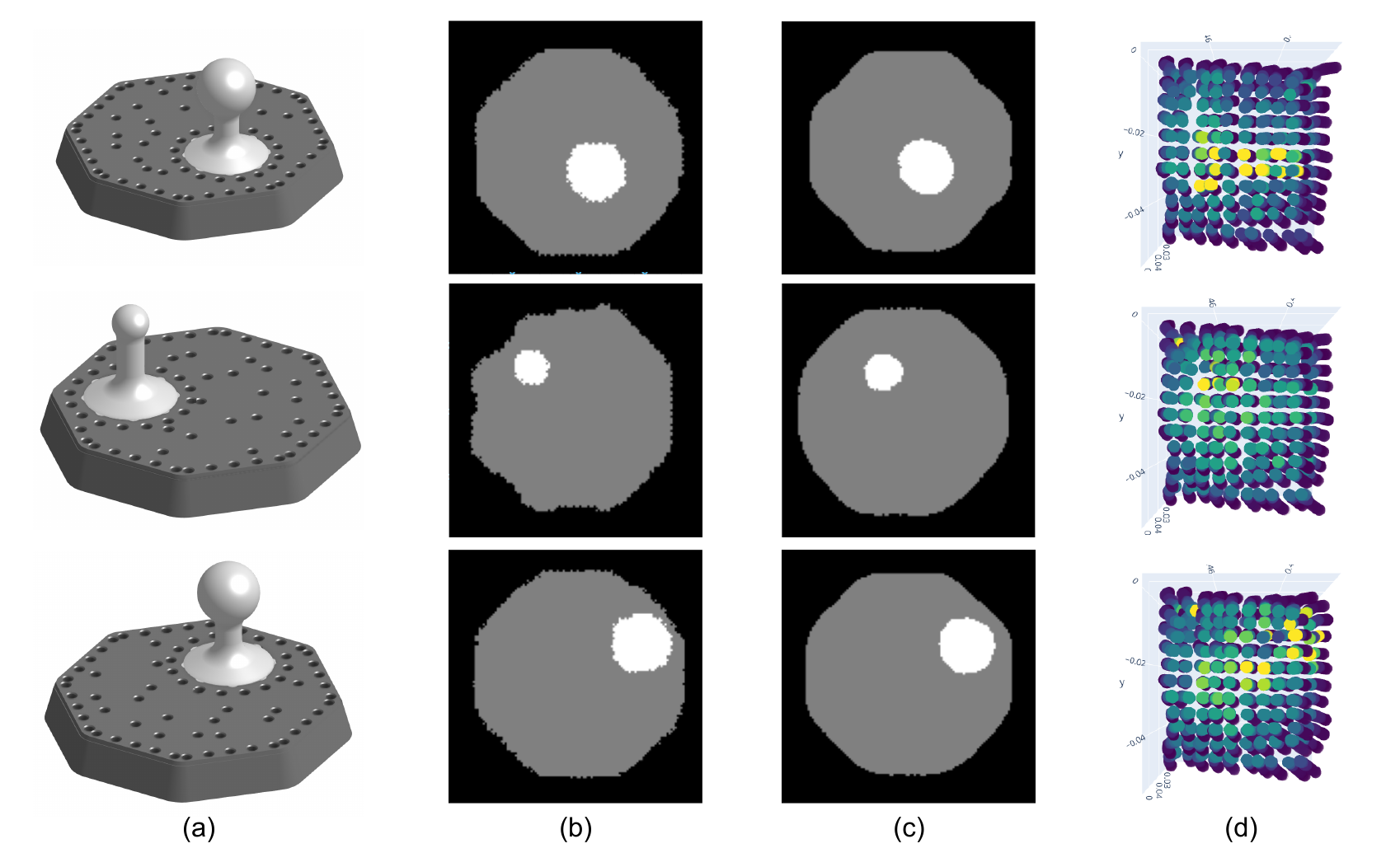}
    \vspace{-0.5em}
    \caption{Tactile imaging with real data. Columns show: (a) 3D CAD design, (b) Ground-truth MRI image slice, (c) Predicted images using our method, (d) Force-map visualization.}
    \label{fig:mri_imaging_vis}
    \vspace{-1em}
\end{figure}

\paragraph{Tactile Imaging:}
In Figure~\ref{fig:mri_imaging_vis} we show several imaging results and their ground truth counterparts; a more extensive demonstration is provided in \cref{sec:imaging_results}. For comparison, we visualize a map of forces from the same data by taking the maximal recorded vertical force across sensors in each location. Evidently, our learning-based approach aggregates the scattered force measurements into a coherent object, as appears in the data, leading to a tactile image that is arguably more interpretable. For interpreting image quality, in addition to the F1 score, we consider two measures that evaluate the lump prediction accuracy: its area in the image (henceforth, size), and the position of its center of mass (CoM).
\begin{table}[h]
\centering
\caption{Lump Size error, Center-of-Mass (CoM) error and F1 score. For prediction methods, we report the standard deviation of the sample mean across 5 random seeds. For the average prediction, we report the standard deviation across samples.}
\vspace{0.3cm}
\label{tab:image_prediction}
\begin{tabular}{lccc}
\toprule
\textbf{Method} & \textbf{Size Error [\%] $\downarrow$} & \textbf{CoM Error [mm] $\downarrow$} & \textbf{F1 Score} [\%] $\uparrow$ \\ 
\midrule
Image Pred. & $23.0 \pm 2.1$ & $2.4 \pm 0.0$ 
 & $74.4 \pm 0.1$\\
Image Pred. ($0.5\times$ Data) & $20.9 \pm 0.9$ & $3.6 \pm 0.1$ & $ 65.3 \pm 0.1$ \\
Image Pred. ($0.25\times$ Data) & $41.6 \pm 3.2 $ & $6.1 \pm 0.4$ & $43.9 \pm 2.6 $ \\
MLP Pred. & $10.5 \pm 0.5$ & $2.9 \pm 0.1$ & - \\
Force Map + Image Pred. & $46.6 \pm 2.8 $ & $5.3 \pm 0.1 $ & $47.3 \pm 1.3$ \\
Average Pred. & $ 51.9 \pm 47.5 $ & $12.1 \pm 4.1 $ & - \\
\hline
\end{tabular}
\end{table}
In Table~\ref{tab:image_prediction}, we provide quantitative results, showing that our method (\textbf{Image Pred.}) achieves $23$\% error in lump size and $2.4$mm error in lump location. In comparison, the average errors for predicting the average lump area and location (\textbf{Average Pred.}) are $52$\% and $12.1$mm, respectively. Therefore, our representation is clearly informative about the lump properties. We also trained a multilayer perceptron (MLP) predictor from our pretrained representation to $x-y$ CoM position and lump size (\textbf{MLP Pred.}). The lump size MLP was more accurate than the image predictor on size prediction, with $10.5$\% error. This shows that the representation contains even more information than is extracted by the image predictor, probably due to insufficient image data. We also evaluated a method that uses the (non-learned) force map as a representation, with the same flow matching image predictor as we used with our representation (\textbf{Force Map + Image Pred.}), see \cref{sec:force_map_baseline} for complete details. This method yielded significantly worse results, which strengthens the importance of our self-supervised learning approach.

As our approach is data-driven, we hypothesize that increasing the amount of data should improve the tactile imaging results. To support this, we repeat the self-supervised pre-training and MRI prediction with $0.5 \times$ and $0.25 \times$ less training data. As can be seen in Table~\ref{tab:image_prediction}, decreasing the amount of training data significantly worsens image prediction performance.

\begin{wraptable}{r}{0.5\textwidth}
\centering
\begin{tabular}{>{\centering\arraybackslash}m{2cm}|
                >{\centering\arraybackslash}m{2cm}|
                >{\centering\arraybackslash}m{2cm}}
 & $\mathbf{0.5\times}$ \textbf{Sup. Data} & $\mathbf{1\times}$ \textbf{Sup. Data} \\ \hline
$\mathbf{0.5\times}$ \textbf{Rep. Data} & $65.3 \pm 0.1$ & $74.5 \pm 1.7$ \\ \hline
$\mathbf{1\times}$ \textbf{Rep. Data} & $72.7 \pm 0.3$ & $74.4 \pm 0.1$ \\ 
\end{tabular}
\caption{F1 score of our proposed tactile imaging approach with varying amounts of unsupervised and supervised data. We report the standard deviation of the sample mean across 3 random seeds.}
\label{tab:datascaling}
\end{wraptable}

\paragraph{Data Scaling:} Compared to non-learning approaches, our data-oriented paradigm
benefits as we train on more data (as shown in \cref{tab:image_prediction}). Still, collecting more labeled data (i.e. tactile measurements, coupled with an MRI scan) is costly. This problem will be even more significant for real human data.
This issue raises a question - \textit{does our approach scale with more unsupervised data only, without more labels?} 

To test this we trained a model with 50\% supervised data and 100\% unsupervised data (and vice versa), the results are shown in \cref{tab:datascaling}. Adding unsupervised data alone resulted in over $11\%$ increase of the F1 score, while adding the supervised data on top of it contributed only an additional $2\%$. In the context of a future study with human data, this result hints that costly labels can be replaced with easier to obtain self-supervised training.

\paragraph{Change Detection:}

To ground our results and highlight the task difficulty, we performed a small-scale human study for the change detection task. We asked $16$ participants to palpate the phantom and detect whether or not we changed the insert to one with a larger lump at the same position (full technical details regarding the human study can be found in Section \ref{app:human_study}). Each participant repeated the experiment up to $6$ times, resulting in $63$ samples. We compared with classification based on our lump size prediction for the same phantoms that were presented to the participants. We obtained a recall of $0.82$ and false-alarm-rate (FAR) of $0.19$, better than the human recall of $0.62$ and FAR of $0.32$. It is remarkable that these results were obtained with the Xela uSkin, for which the spatial density of sensors is almost two orders of magnitude smaller than the human finger~\citep{corniani2020tactile}. To further appreciate the task's difficulty, we provide a video in the supplementary material that visualizes how the lump moves inside the hydrogel when touched.

\paragraph{Shell Classification:}
To further demonstrate the information content in our learned representation, we show in \cref{sec:shell_classification} of the supplementary material that it can be used to reliably predict the phantom shell ID (out of 4 possible shells in our data) -- information that is not available in the MRI data. Thus, the representation is informative enough to capture small artifacts in the manufacturing process (all shells are manufactured in the same way), effectively distinguishing between the shells with an accuracy of $99.6 \pm 0.7$.

%% file: sections/discussion.tex
\section{Discussion} \label{sec:discussion}

Our results indicate that with enough data, a neural network trained to process tactile measurements yields a representation that is informative about the palpated body. Importantly, the performance of our method improves with additional data, without changing the hardware -- unlike conventional tactile imaging methods that are not learning-based. How much, and what kind of, data is required to scale our approach to be clinically relevant? 

Tactile measurements depend on the body being palpated, the sensor type, and the sensor movement. Relating to findings in other modalities such as vision and language, we predict that collecting massive tactile data from multiple sensors on general objects (not necessarily soft phantoms) may lead to foundation models for touch processing that can be fine-tuned to specific palpation tasks. The large dataset collected here can contribute to a collective data collection effort~\citep{o2024open,bell2023sharing}. One interesting question is whether language, which has played a key role in foundation models for vision and robotics, is also important for touch (cf. the manual palpation instructions mentioned in \cref{sec:intro}).

Our data collection protocol requires a robot both for automation, but also for pose estimation. Extending our results to a human moving the sensor -- a likely application, requires adjustments of the palpation trajectories we record to be more human-like or even apply active sensing~\citep{scimeca2022action}, while pose estimation has standard solutions such as fiducial markers~\citep{fiala2005artag}.

Relating to our imaging results, MRI of human breast tissue reveals intricate details of small structures like ducts and blood vessels that do not exist in our fabricated models, making them significantly more complex to predict. In addition, the common use of contrast agent injections in breast scans to enhance the visibility of cancerous tumors is not accounted for in our current models. Thus, we cannot immediately deduce that our results will generalize to real human data. Nevertheless, as there is a correlation between abnormalities in breast tissue and their tactile sensing, we see promise in further investigating this direction. Moreover, although a vast body of work has explored deep learning for MRI \citep{heckel2024deep}, to the best of our knowledge, learning the mapping from sensory data to MR images has not been explored yet, and our work initiates this new line of research.

Finally, in addition to the technological challenges in artificial palpation, there are clinical and sociological challenges. While several clinical trials have been performed with conventional tactile imaging~\citep{NCT06643767}, we do not yet have information about the sensitivity and specificity of learning based approaches, or palpation using low-cost sensors. On a positive note, the survey by \citet{jenkinson2023acceptability} showed a generally positive reaction to automated breast cancer screening.

\section{Acknowledgements}
This work received funding from the European Union (ERC, Bayes-RL, Project Number
101041250). Views and opinions expressed are however those of the authors only and do not necessarily reflect those of the European Union or the European Research Council Executive Agency.
Neither the European Union nor the granting authority can be held responsible for them. E.S. is a Horev Fellow and acknowledges funding support from the Technion's Leaders in Science and Technology program and Alon Fellowship. We are grateful for the support of the May-Dahl-Blum Technion Human MRI research center staff and services.
\newpage

%% file: sections/data_collection.tex
\section{Robotic Data Collection Setup} \label{sec:datacollection}

While the simulation setup is useful for quick development iterations, it does not mimic the real-world, and the learned representations will fail to generalize outside the simulation.
To test our approach on a real tactile sensor we designed a setup to mimic a simplistic palpation examination.

\subsection{Xela uSkin and Digit Comparison} \label{sec:digit_vs_xela}

We use a single Xela uSkin \citep{xelauskin} 
which has 30 force sensors, each measuring forces in $x$, $y$, and $z$ directions.

\begin{figure}[htbp]
\centering
\begin{minipage}[t]{0.23\textwidth}
    \centering
    \includegraphics[width=\linewidth]{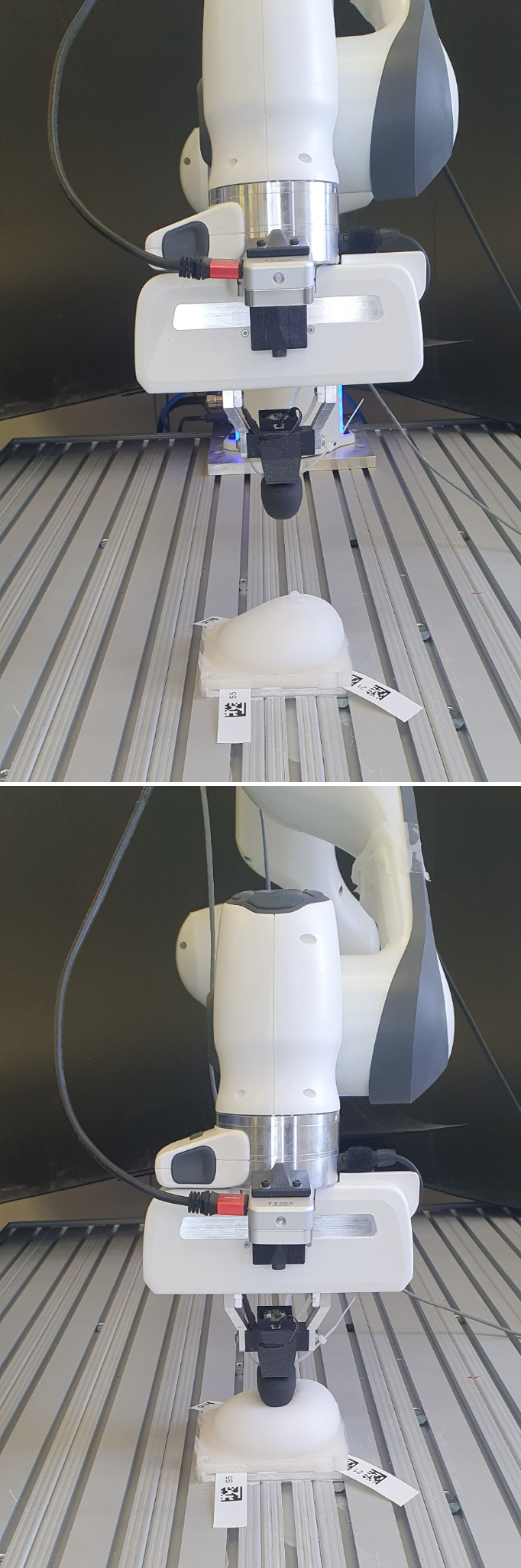}
    \caption*{(a)}
\end{minipage}
\hspace{0.1em}
\begin{minipage}[t]{0.23\textwidth}
    \centering
    \includegraphics[width=\linewidth]{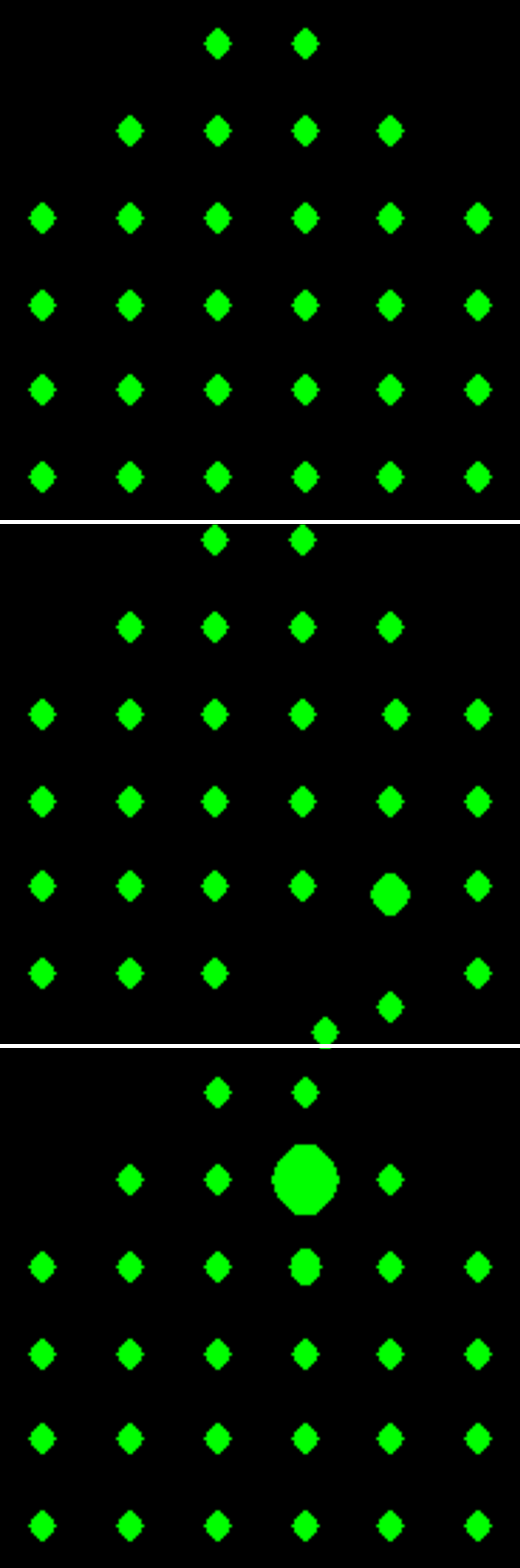}
    \caption*{(b)}
\end{minipage}
\hspace{0.1em}
\begin{minipage}[t]{0.23\textwidth}
    \centering
    \includegraphics[width=\linewidth]{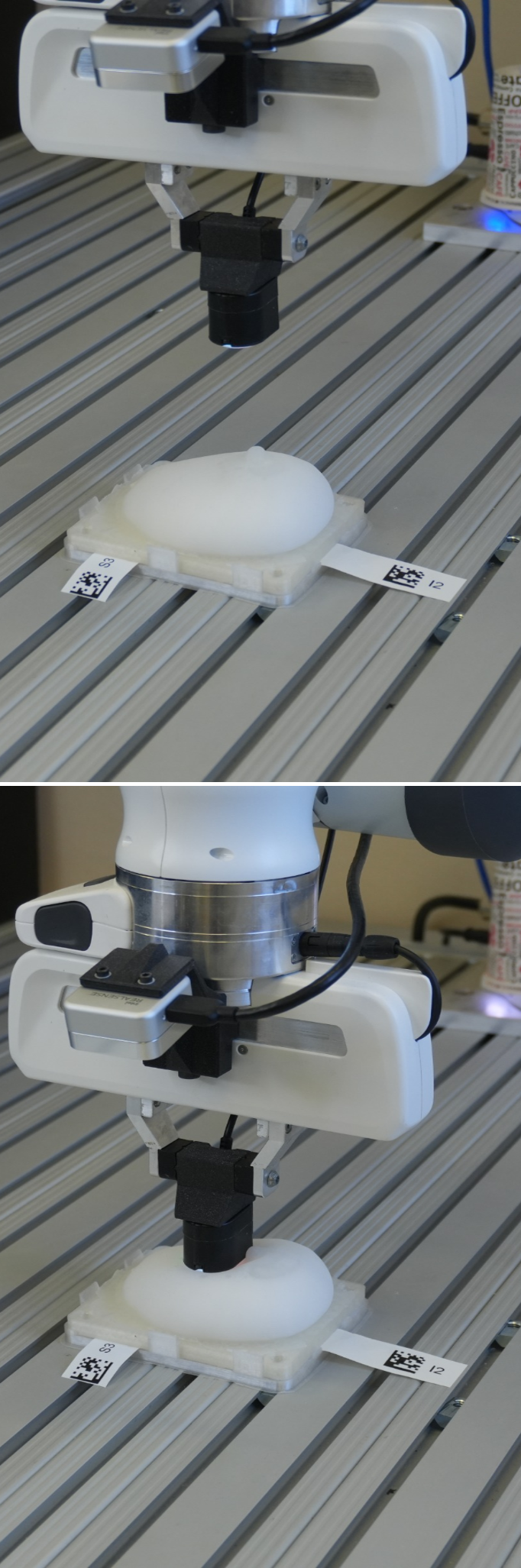}
    \caption*{(c)}
\end{minipage}
\hspace{0.1em}
\begin{minipage}[t]{0.23\textwidth}
    \centering
    \includegraphics[width=\linewidth]{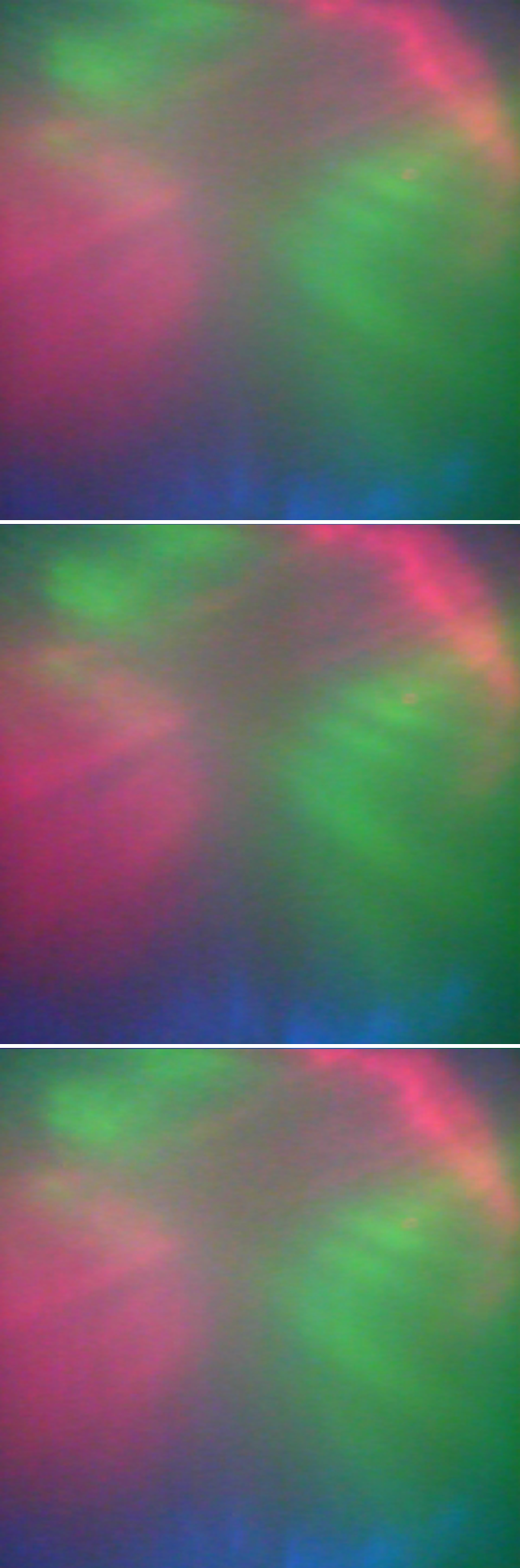}
    \caption*{(d)}
\end{minipage}

\caption{
Comparison of tactile perception pipelines. (a), (b): Xela system — robot and tactile sensor views. (b) shows captures from the Xela visualizer. The top image depicts a state of no touching, the middle one touching a phantom without a lump, and the bottom one touching a lump. (c), (d): Digit system — robot and tactile sensor views. (d) Images obtained by the Digit sensor. The sensor positions are the same as in (b).
}
\label{fig:xela_vs_digit_comparison}
\end{figure}

In our work, we have not made strong assumptions on the tactile sensor, and in theory, our learning approach can be used with other sensors. We have tested the behavior of a Digit sensor \citep{lambeta2020digit}. 
To test the behavior of Digit compared to the uSkin, we performed a simple experiment. We first touched a phantom containing an insert without a lump, and then touched an insert with a lump exactly on top of the lump. We repeated this test with the Digit and uSkin, and the results are reported in \cref{fig:xela_vs_digit_comparison}. As can be seen in the Figure, when touching the samples with Xela sensor, we get significant measurements when touching the shell, while the digit doesn't seem to be affected by it. We tried to measure various objects with the digit sensor and saw that it is very good at detecting surface patterns and is responsive to hard surfaces, but is very lacking in sensitivity to soft objects and depth changes. In an experiment we conducted earlier, we tried the Xela vs. the digit in a previous generation of inserts. We easily classified the touching lump / no lump with the Xela and completely failed with the digit. 

\subsection{Breast Phantoms}

\begin{figure}[htbp]
     \centering
     \begin{subfigure}[b]{0.39\textwidth}
         \centering
         \includegraphics[width=\textwidth]{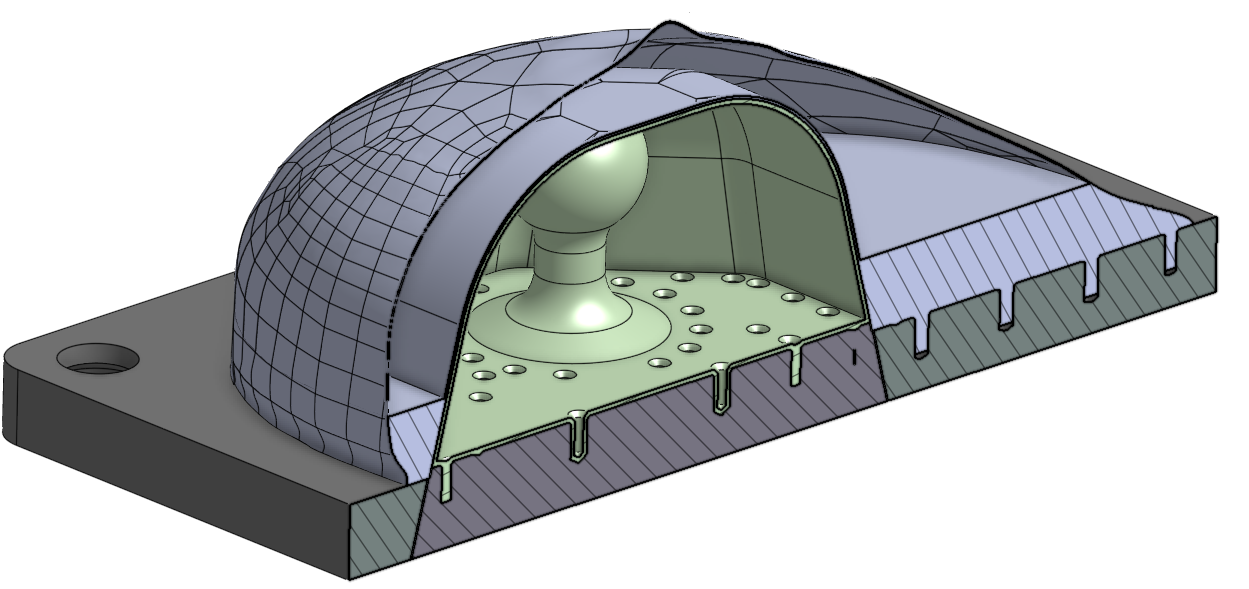}
         \caption{}
         \label{fig:assy_section_view}
     \end{subfigure}
     \hfill
     \begin{subfigure}[b]{0.30\textwidth}
         \centering
         \includegraphics[width=\textwidth]{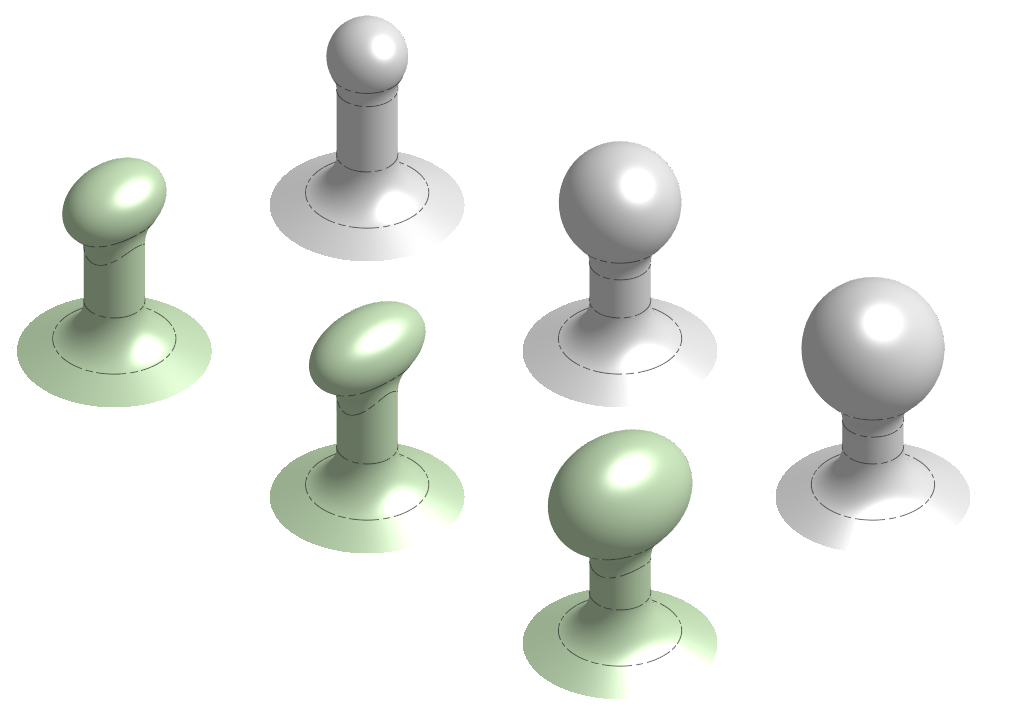}
         \caption{}
         \label{fig:six_lumps}
     \end{subfigure}
     \hfill
     \begin{subfigure}[b]{0.29\textwidth}
         \centering
         \includegraphics[width=\textwidth]{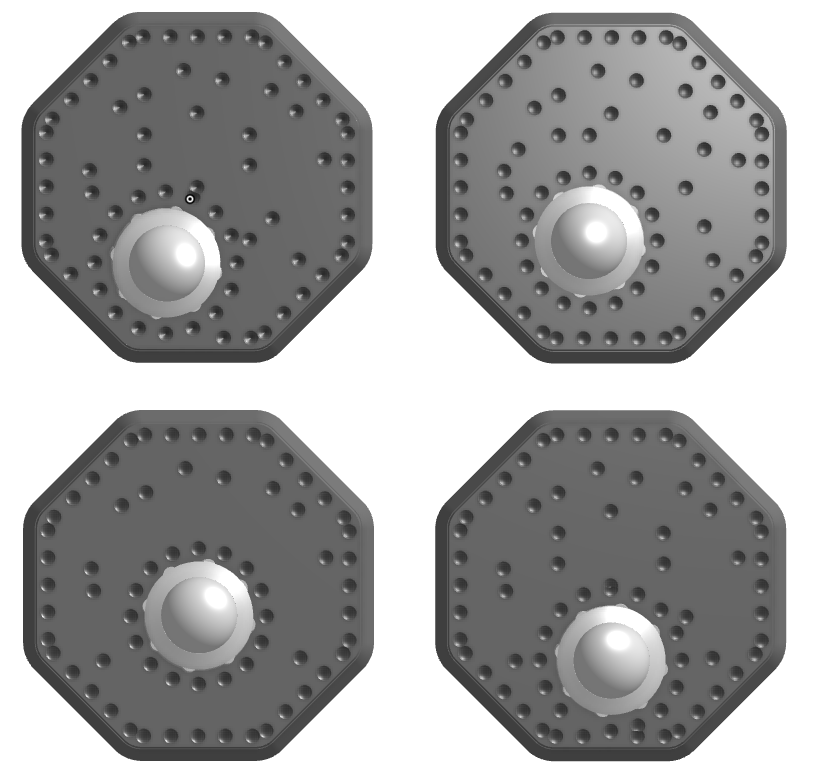}
         \caption{}
         \label{fig:four_bases}
          \hfill
     \end{subfigure}
        \caption{In (a): Section view of the shell and insert assembly. In green, insert silicone skin. In blue, shell silicone skin. In dark gray, PLA bases for the insert and shell. Voids are filled with slime. Note that holes in the bases are filled with silicone for anchoring. (b) Lump sizes: In white from furthest to closest, 8, 12, and 14mm diameter spheres. In Green ellipsoids with long/short axes of 12/8, 14/8, and 16/12mm. In (c), four insert bases, each with a 14mm lump at a different location. Other lumps (not shown) can be inserted at the same locations.}
        \label{fig:phantom_models_appendix}
\end{figure}

For the breast phantoms, we aimed for a modular design, which will allow for effective data collection. Our phantoms are built from two parts - a shell and an insert. Both shell and insert are made using a thin outer layer of silicone (Smooth On Dragonskin FX) filled with "slime" - a gel matrix made of polyvinyl acetate (PVAc), borax, and water. We shall refer to the outer layer as skin. The skin is attached to a base plate made of polylactic acid (PLA) plastic using pre-made holes in the base plate.
Inside the insert, we have the lump, which is made of the same Dragonskin FX silicone as the outer layer. See \cref{fig:assy_section_view} for a section view of the assembly.
The insert has an octagon-shaped base plate that fits into the shell's base plate, this is so we can have eight configurations for each insert, lessening the amount of inserts needed. We created 24 inserts in the following configuration: 4 lump locations as seen in \cref{fig:four_bases}, with 3 spherical and 3 ellipsoidal lump sizes as seen in \cref{fig:six_lumps}.

\subsection{Phantoms Fabrication} 
\begin{figure}[htbp]
     \centering
     \begin{subfigure}[b]{0.5\textwidth}
         \centering
         \includegraphics[width=\textwidth]{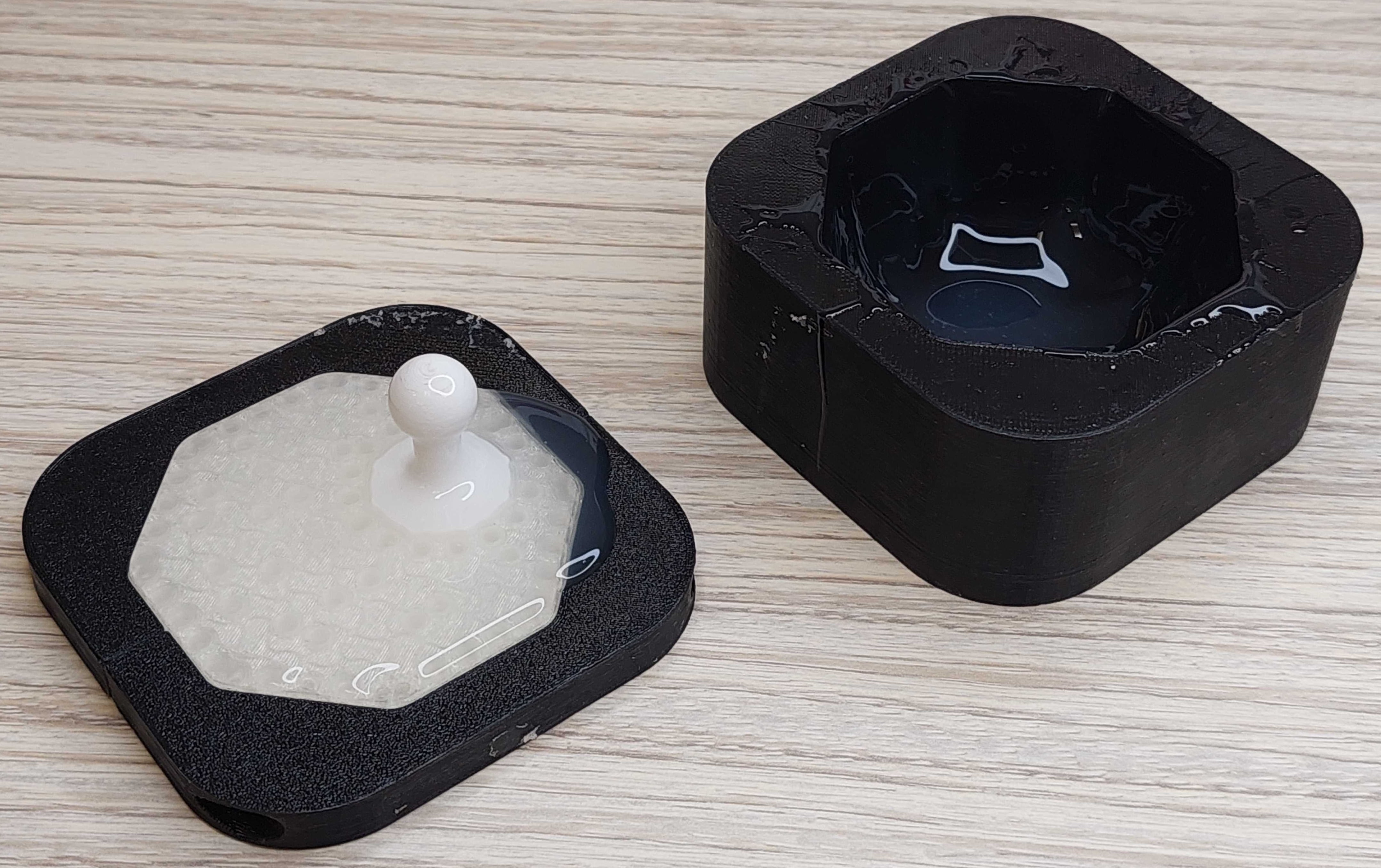}
         \caption{}
         \label{fig:fab1}
     \end{subfigure}
     \hfill
     \begin{subfigure}[b]{0.45\textwidth}
         \centering
         \includegraphics[width=\textwidth]{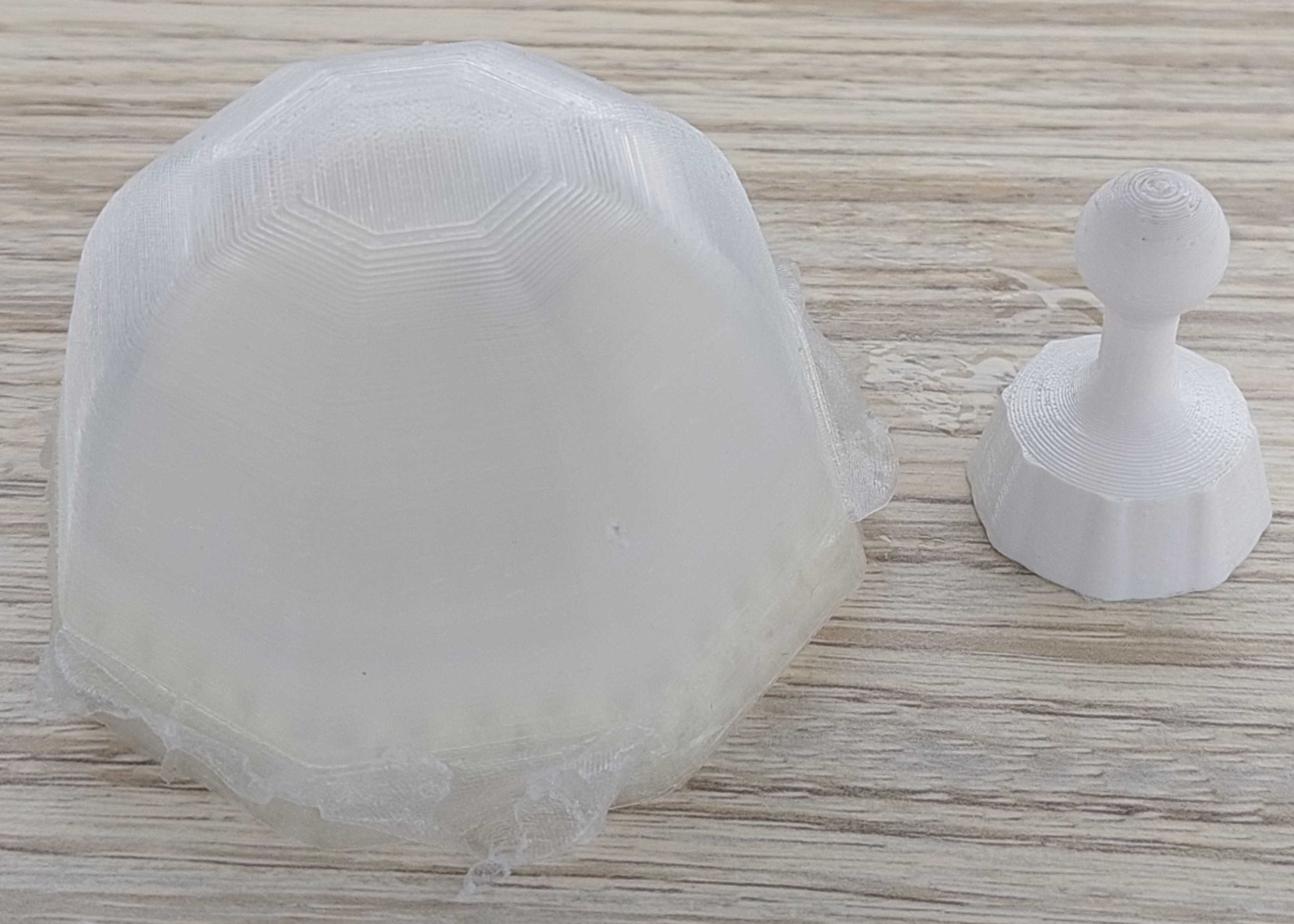}
         \caption{}
         \label{fig:fab2}
     \end{subfigure}
          \hfill
    \begin{subfigure}[b]{0.35\textwidth}
         \centering
         \includegraphics[width=\textwidth]{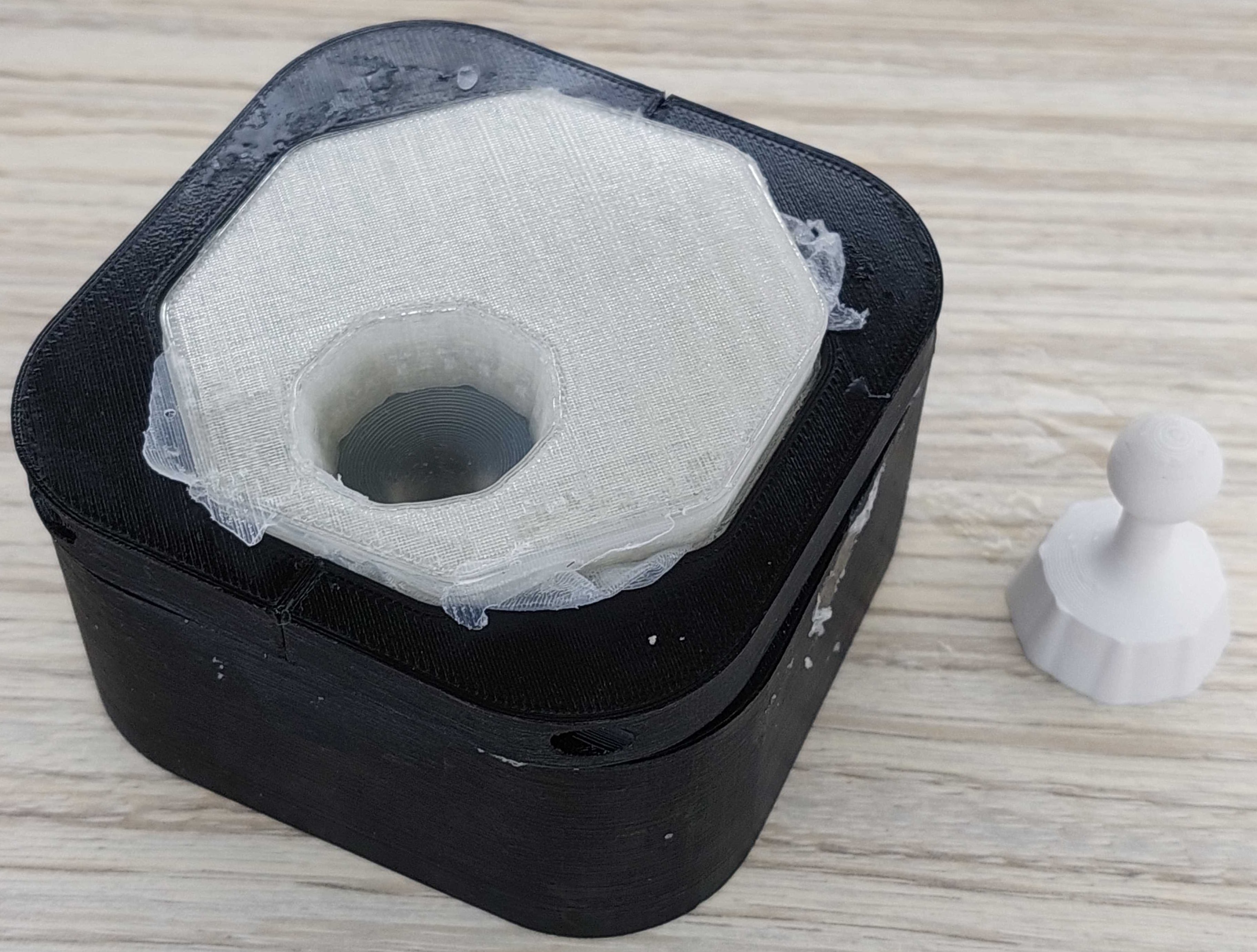}
         \caption{}
         \label{fig:fab3}
     \end{subfigure}
          \hfill
     \begin{subfigure}[b]{0.3\textwidth}
         \centering
         \includegraphics[width=\textwidth]{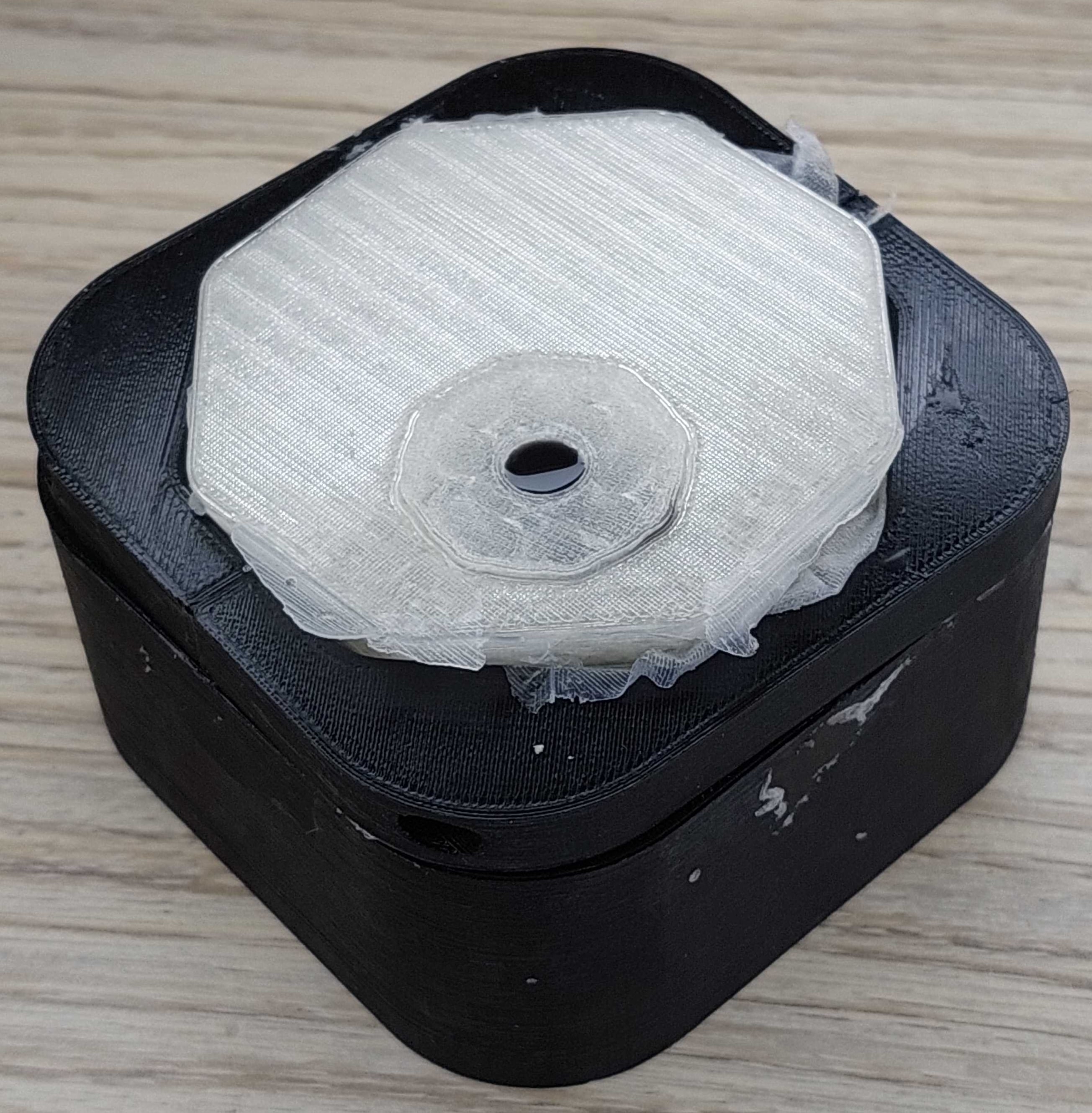}
         \caption{}
         \label{fig:fab4}
     \end{subfigure}
     \hfill
     \begin{subfigure}[b]{0.33\textwidth}
         \centering
         \includegraphics[width=\textwidth]{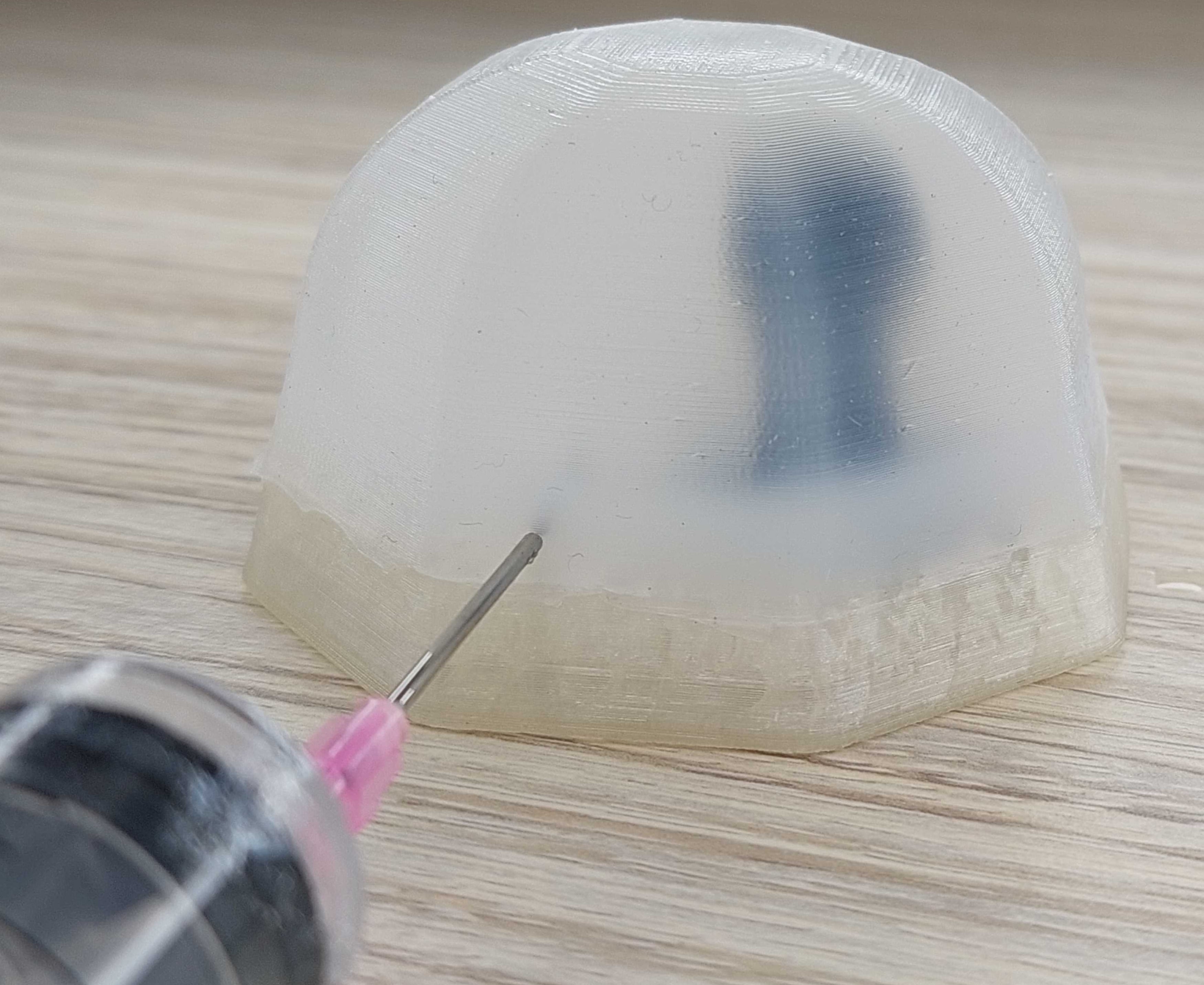}
         \caption{}
         \label{fig:fab5}
     \end{subfigure}
        \caption{Fabrication process for insert. (a) Silicone on the mold parts is fully covered. (b) Parts after removal from the mold. (c) using mold as support. (d) Hole filled with silicone. The silicone is colored black for illustration purposes. (e) Injection of slime.}
        \label{fig:phantom_fabrication}
\end{figure}

The skin for both the insert, and shell is constructed much like hollow chocolate bunnies or Easter eggs: we pour a small amount of silicone into a multipart mold and brush all the surfaces (\cref{fig:fab1}), next we close the mold and rotate it until it sets. When we open the mold, we have a hollow shell or insert.
During the molding process the silicone enters holes in the base, which serve as anchor points for the silicone as seen in \cref{fig:assy_section_view}. For the shell, the process with silicone ends at this stage. In the case of the insert, we have to remove the lump mold (\cref{fig:fab2,fig:fab3}), pour silicone into the lump skin, and close the hole in the base with a plug (\cref{fig:fab4}).

Following the silicone stage, we then inject the slime using a needle through a thick part of the phantom (\cref{fig:fab5}); this works for both shell and insert. This concludes fabrication of the phantom.

\subsection{Transparent Insert}

\begin{figure}[htbp]
     \centering
     \includegraphics[width=\textwidth]{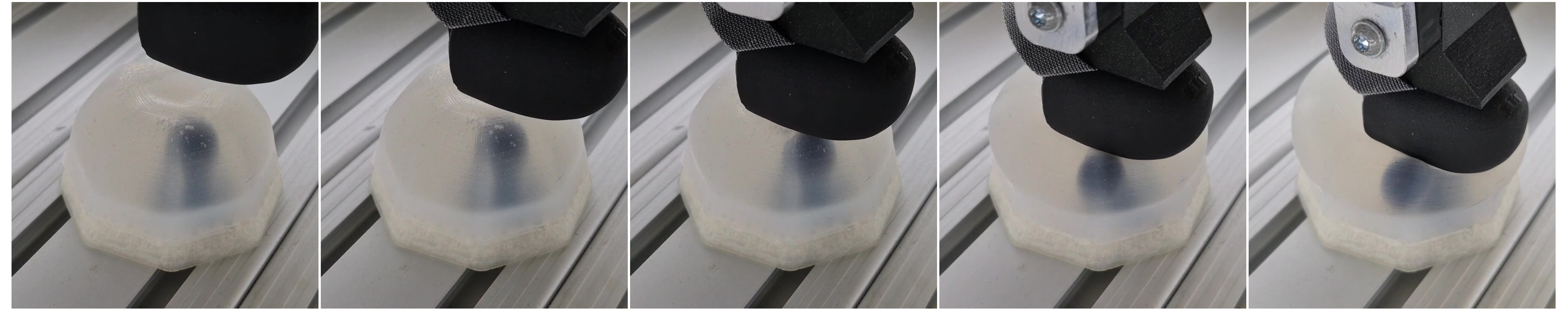}
    \caption{Poke trajectory on a special transparent insert.}
    \label{fig:traninsert_robot}
\end{figure}

To showcase the task's difficulty, we manufactured a single transparent insert so that the lump will be visible during handling. As can be seen in \cref{fig:traninsert_robot}, the lump moves when touched. This is a realistic phenomenon of the 3D structure that would appear in real tissue.    

\subsection{Data Collection Setup Design}

\begin{figure}[htbp]
    \centering
    \includegraphics[width=\linewidth]{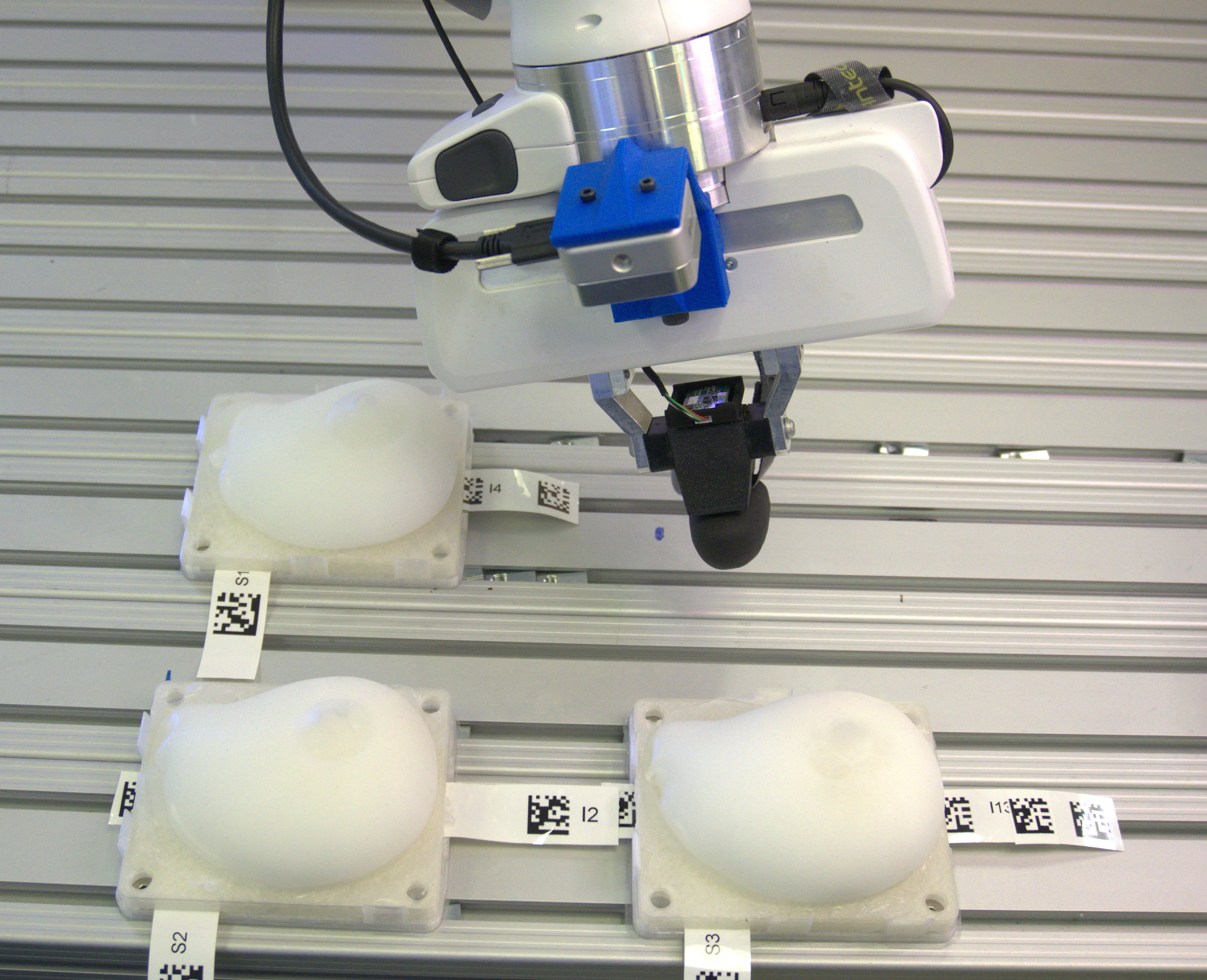}
    \caption{Our automatic robotic data collection setup consists of our breast phantom (including visible QR codes), a Franka Emika Panda robotic arm with a mounted Xela uSkin tactile sensor, and an RGB camera.}
    \label{fig:robotic_setup}
\end{figure}

To collect data in a consistent and repeatable manner, we utilize a Franka Emika Panda robotic arm with 7-DoF. The complete setup, which can be seen in \cref{fig:robotic_setup}, consists of the robotic manipulator, the tactile sensor held as the end effector, a camera attached to the manipulator, and the phantom. Both are attached/held using 3d printed adapters. We controlled the panda using panda-py \citep{elsner2023taming}. For the actual data sampling, we edited the force controller to be a force-position controller in order to press the phantom with a specific force at a specific point. In order to be organized and automatic as possible, we used the camera to scan data matrices (similar to QR codes) that are attached to the shells and inserts to automatically record the used items and the insert orientation. Data was recorded for trajectories at 110 points on the phantom. We recorded Xela data at ~$85$[$Hz$], robot data  at $100$[$Hz$] and RGB images from the camera at $10$[$Hz$]. We collected the data with a fixed force and fixed sensor orientation.

\subsection{Data Collection Technical Details}
The robotic manipulator first scans the attached data matrices (similar to QR codes) to determine the labels of the phantom and the insert, and also to obtain the orientation of the insert inside the phantom. The robot goes to each $x,y$ position ($110$ trajectories) and starts to descend in a straight line in $z$ by $2$[mm] steps until we sense a significant force with the Xela sensor (the sensor output is noisy), and then backtracks $2$[mm]. after that we use a force/position controller that controls the force in z and the position in $x,y$ of the end effector for $5$ seconds, during that time we record all the Xela measurements (~$85$[Hz]), data from the robot ($100$[Hz] $q$, $dq$, calculated force, $tau\_J$,$O\_T\_EE$ ) and RGB images from the realsense camera ($10$[Hz] currently not in use). We considered using the depth capability of the RealSense camera, but it had too much noise in these settings. We add the relevant timestamps to all the data. Robot data consists of the libfranka variables q, dq - joint angles and angular velocity, calculated force - force and torque at the end effector calculated using the Jacobian, $\tau_J$ - Measured link-side joint torque sensor signals, and $O\_T\_EE$ -  Measured end effector pose in base frame.
After testing a number of positions/sensor angles/forces, we decided that the best scan for our needs, which takes ~$20$ minutes, will consist of $110$ positions on the phantom with a single fixed force $3.8$[N] and a single sensor orientation.

%% file: sections/appendix.tex
\section{{PalpationSim}}
\subsection{Simulation Visualization}

\begin{figure}[htbp]
    \centering

    \begin{subfigure}[b]{0.19\textwidth}
        \includegraphics[width=\linewidth]{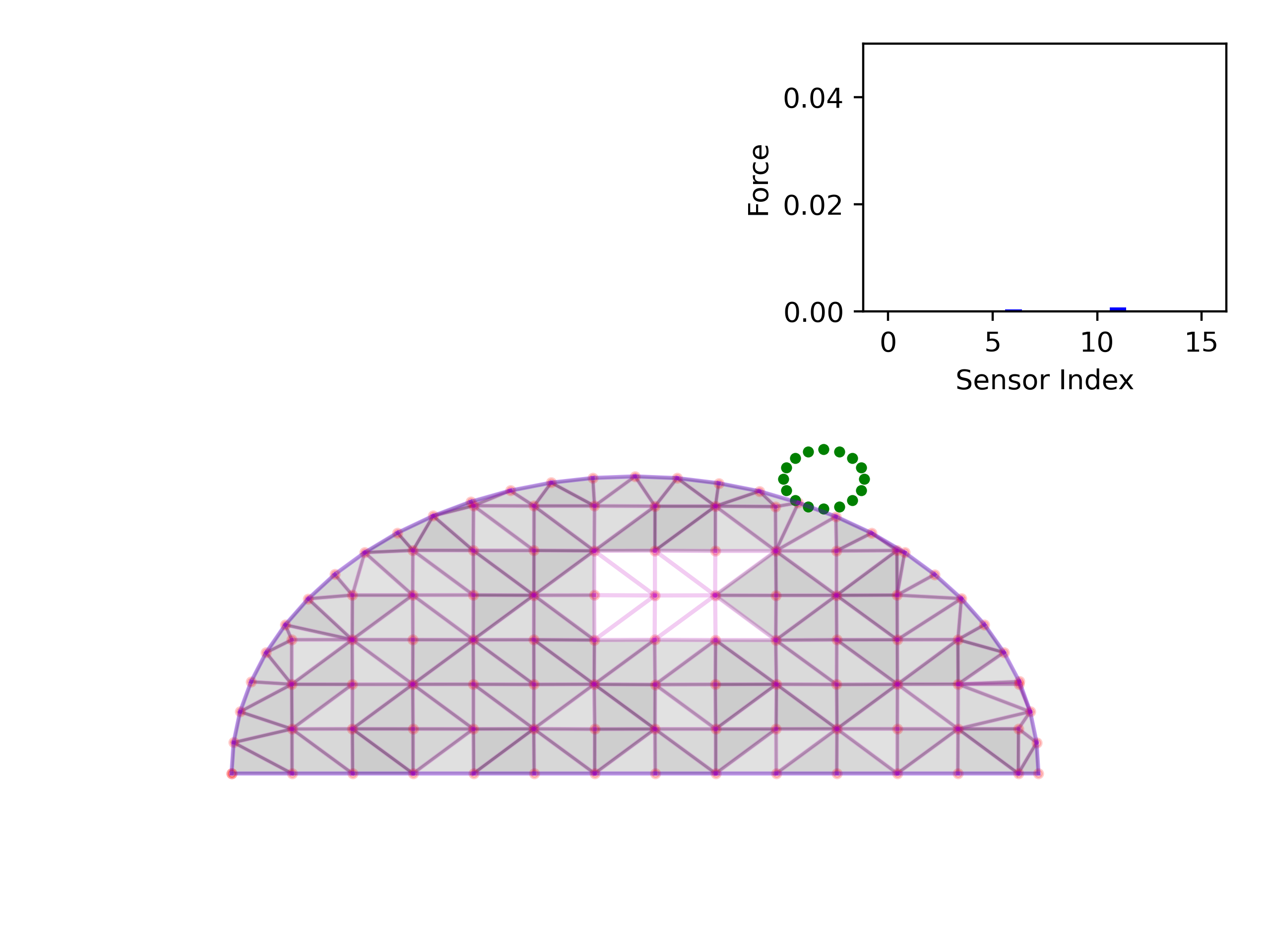}
    \end{subfigure}
    \begin{subfigure}[b]{0.19\textwidth}
        \includegraphics[width=\linewidth]{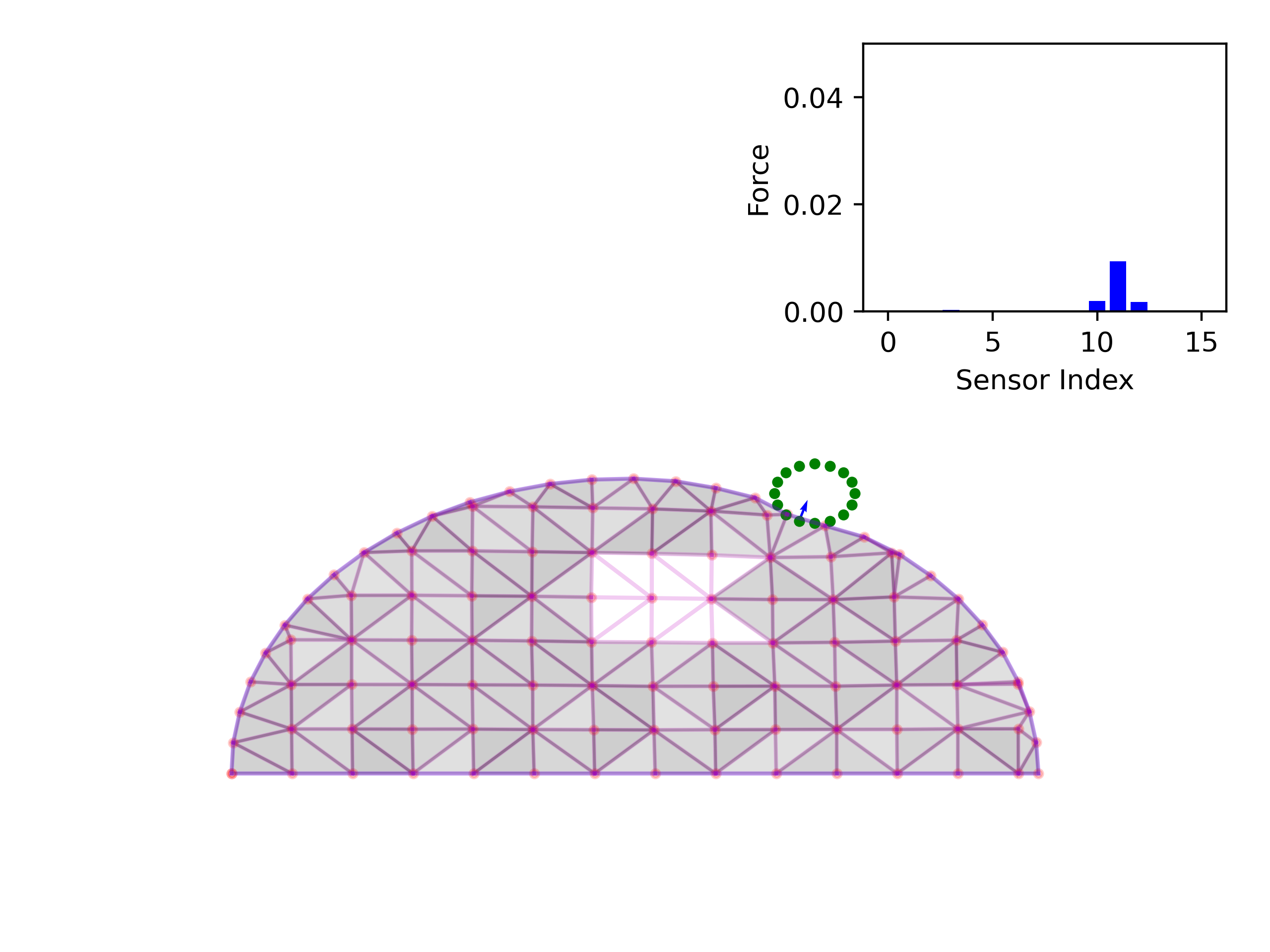}
    \end{subfigure}
    \begin{subfigure}[b]{0.19\textwidth}
        \includegraphics[width=\linewidth]{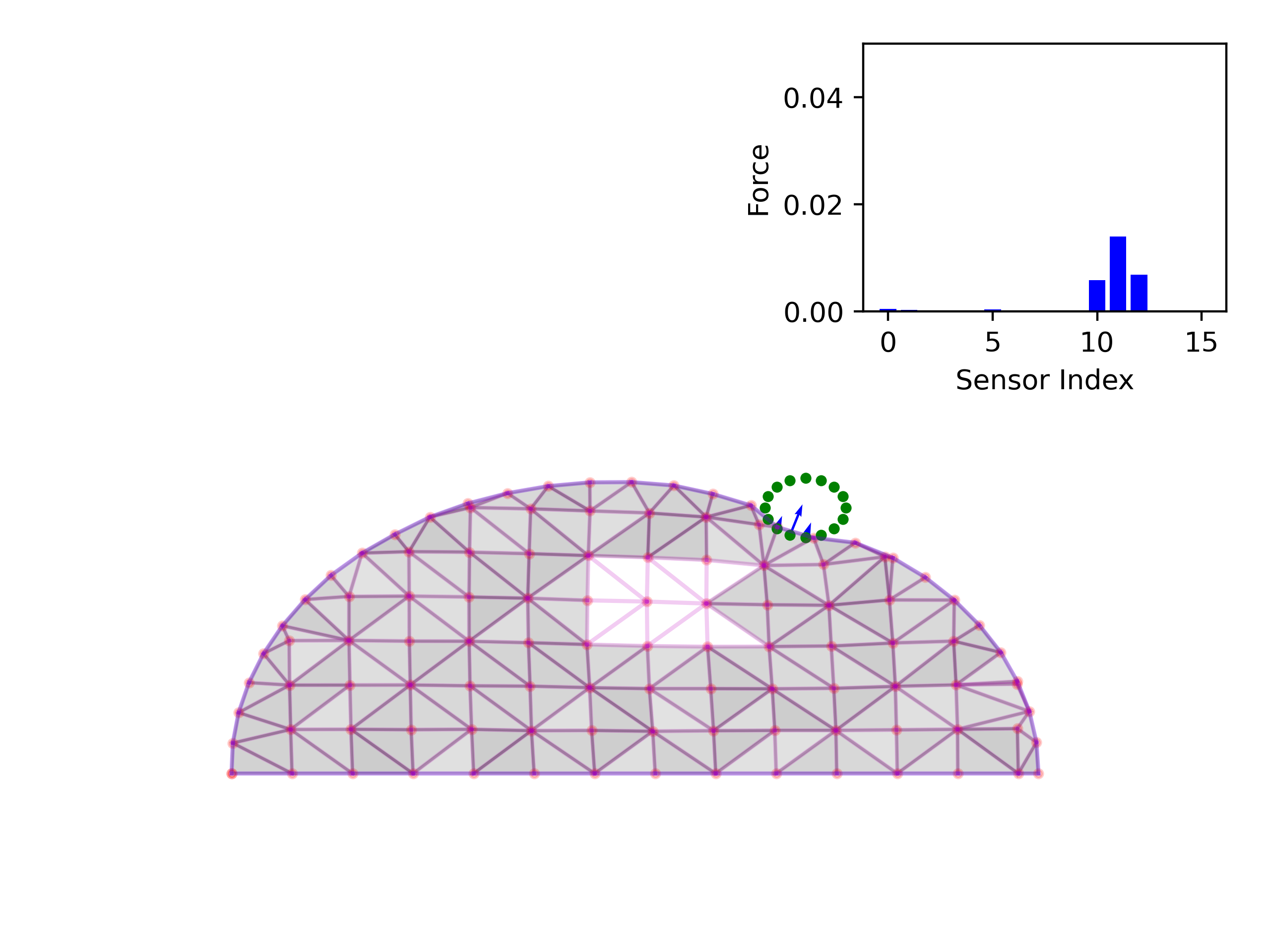}
    \end{subfigure}
    \begin{subfigure}[b]{0.19\textwidth}
        \includegraphics[width=\linewidth]{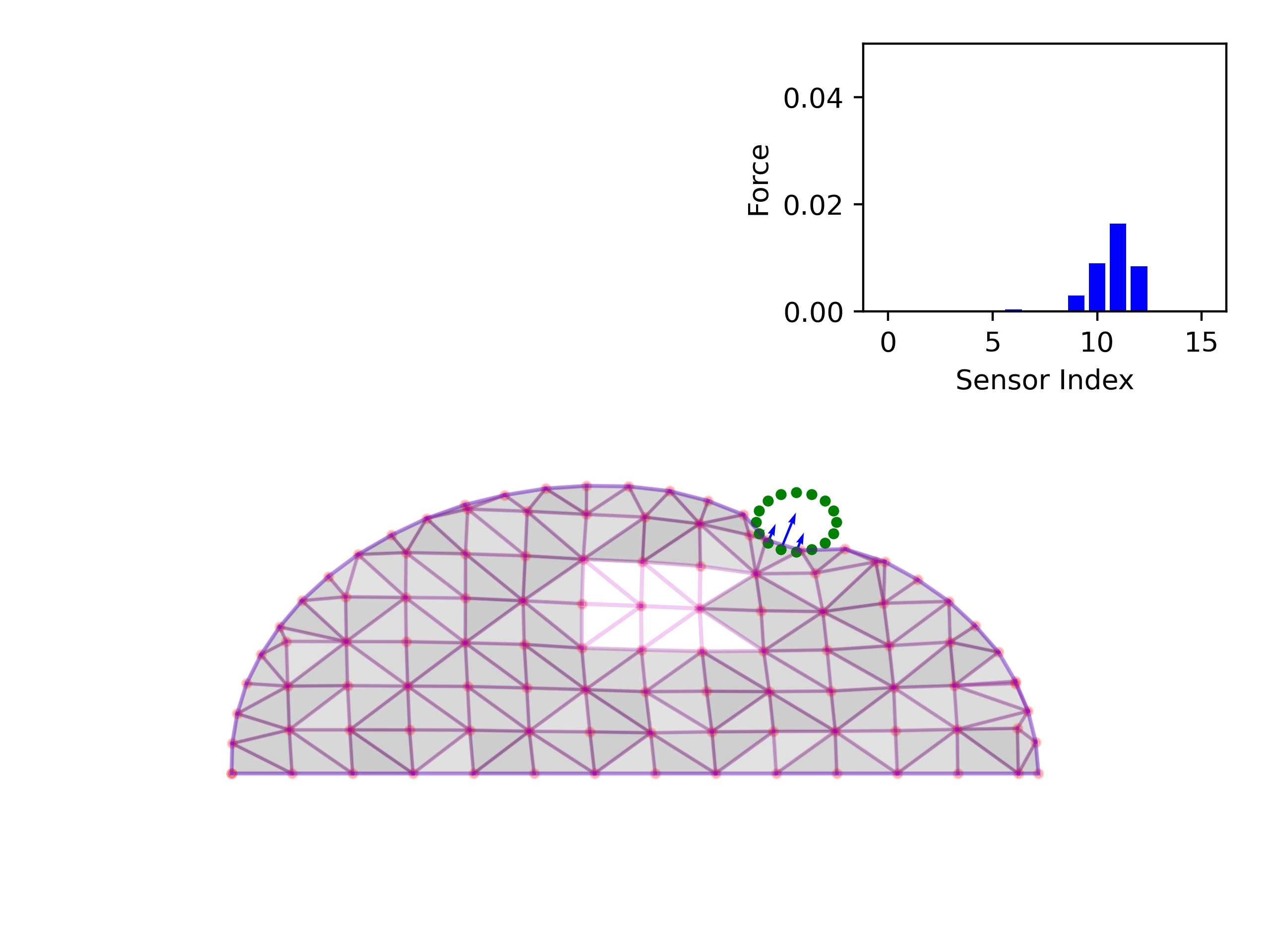}
    \end{subfigure}
    \begin{subfigure}[b]{0.19\textwidth}
        \includegraphics[width=\linewidth]{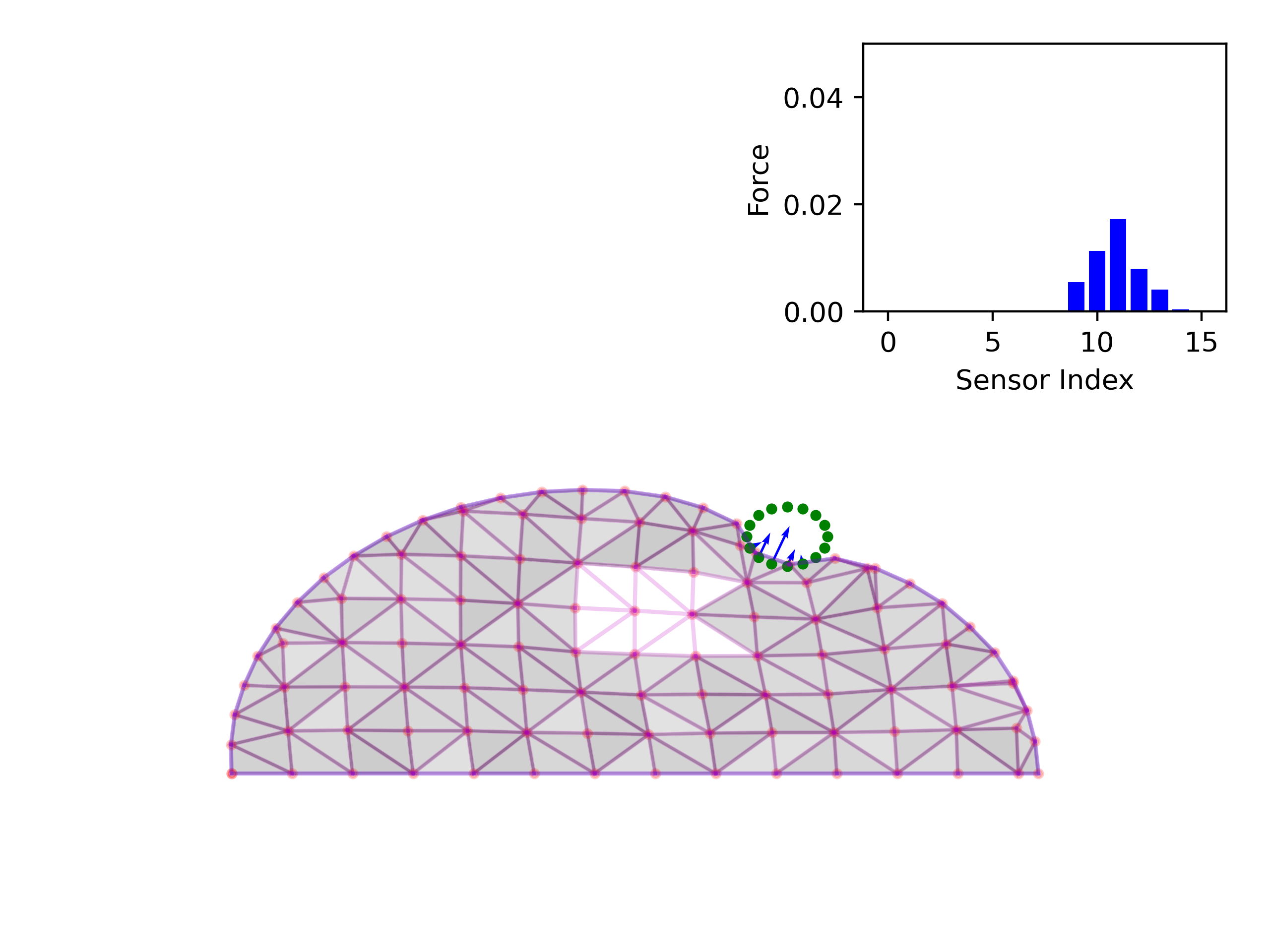}
    \end{subfigure}

    \vspace{0.5em}

    \begin{subfigure}[b]{0.19\textwidth}
        \includegraphics[width=\linewidth]{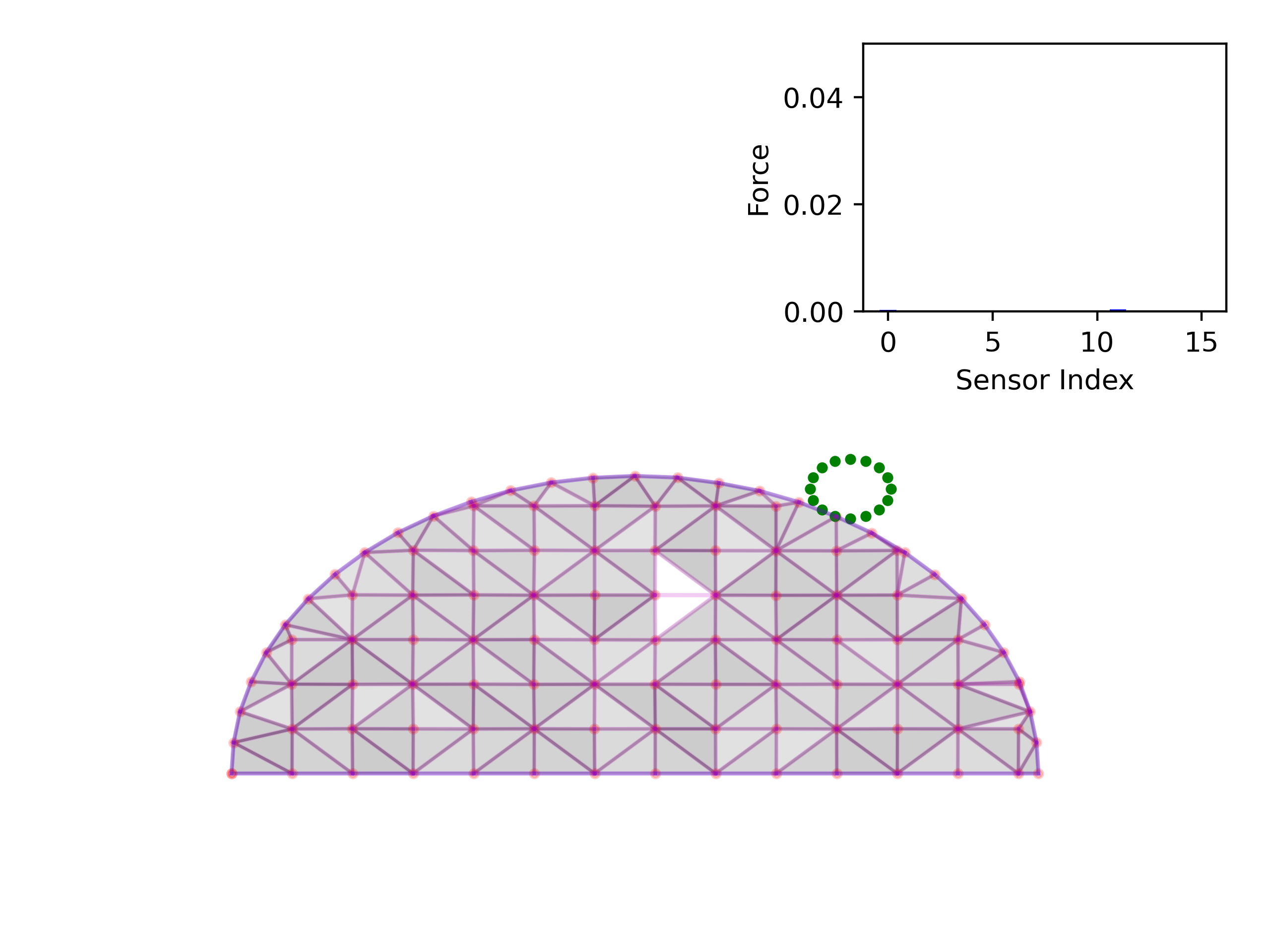}
    \end{subfigure}
    \begin{subfigure}[b]{0.19\textwidth}
        \includegraphics[width=\linewidth]{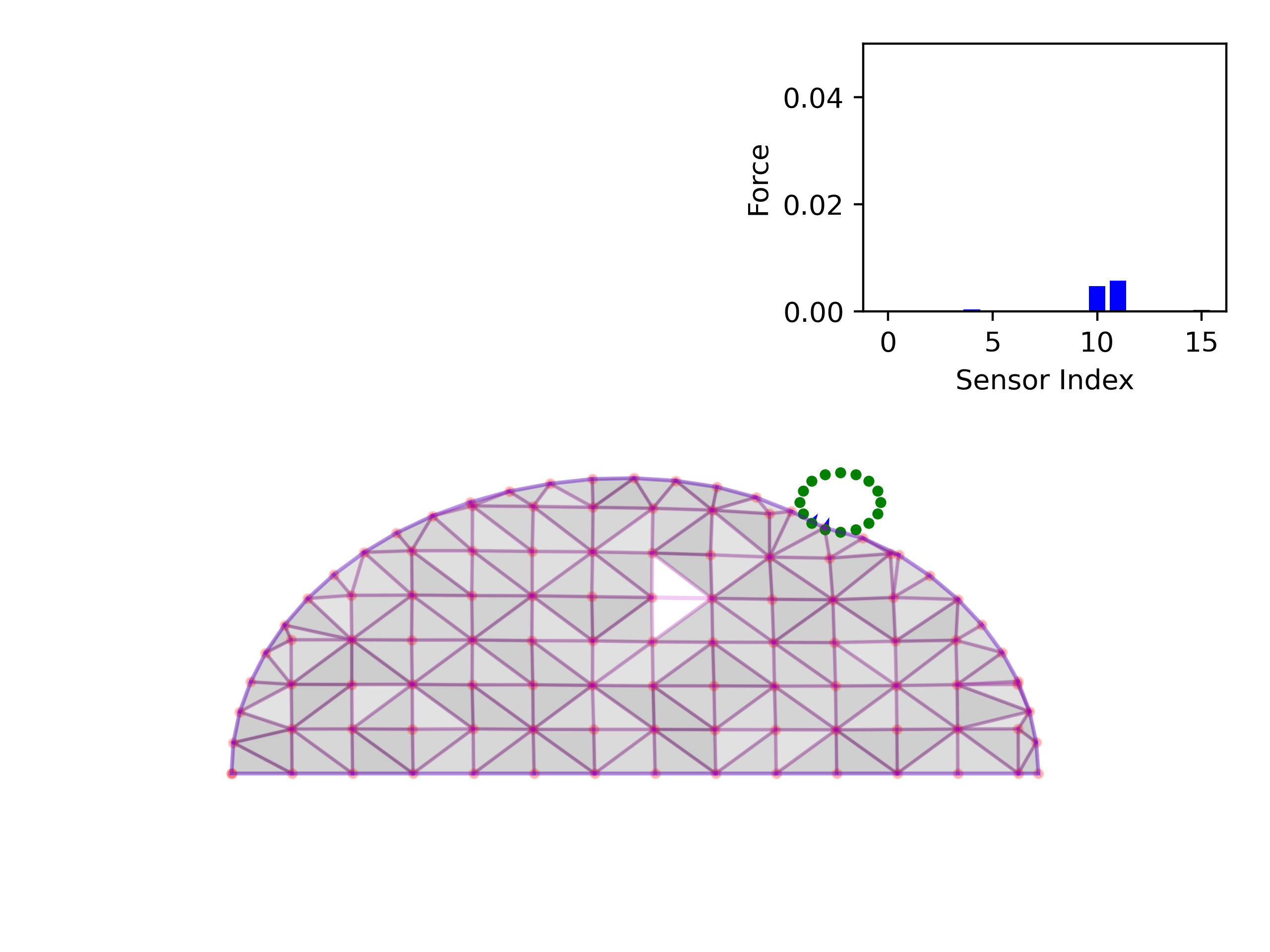}
    \end{subfigure}
    \begin{subfigure}[b]{0.19\textwidth}
        \includegraphics[width=\linewidth]{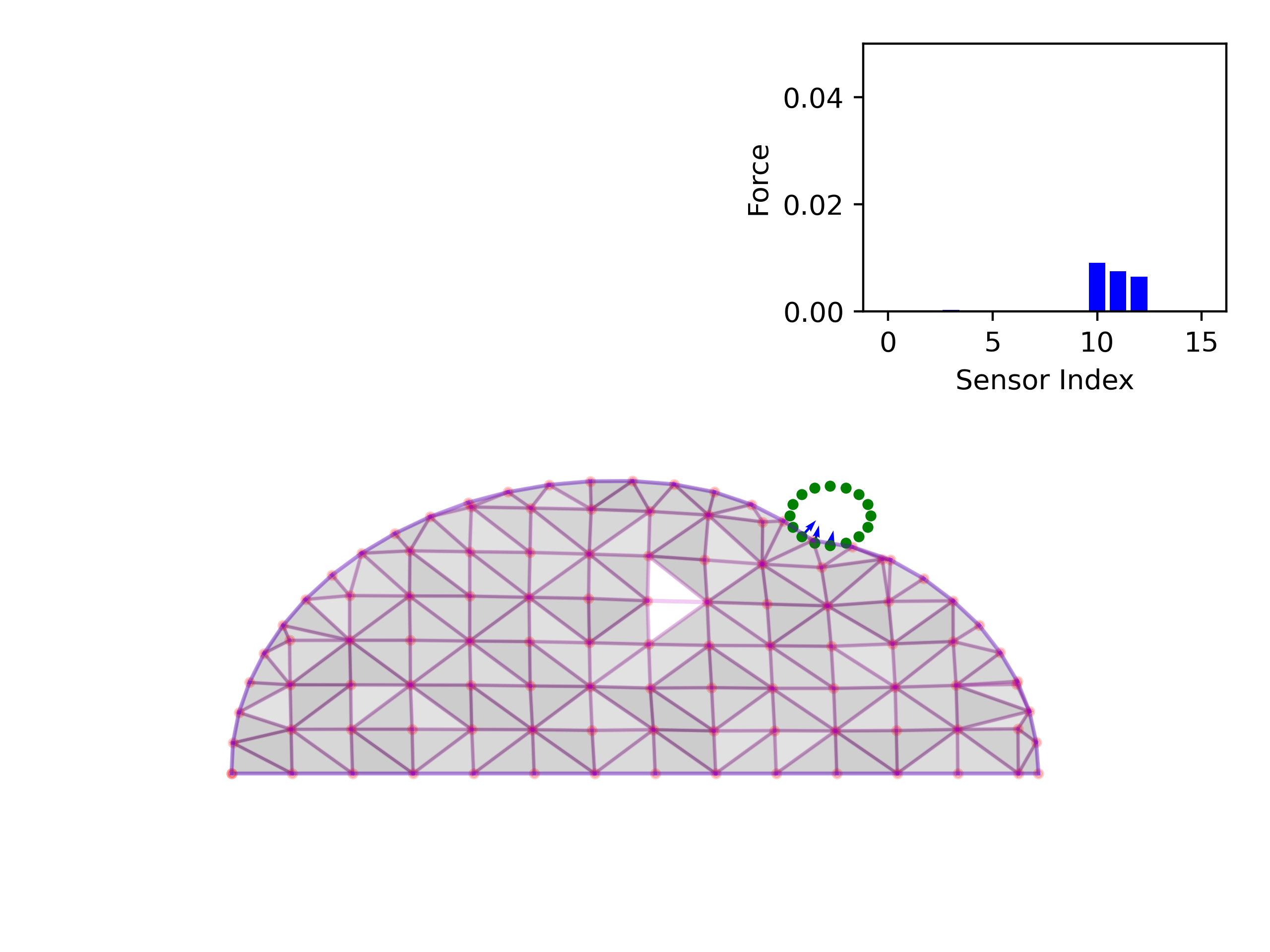}
    \end{subfigure}
    \begin{subfigure}[b]{0.19\textwidth}
        \includegraphics[width=\linewidth]{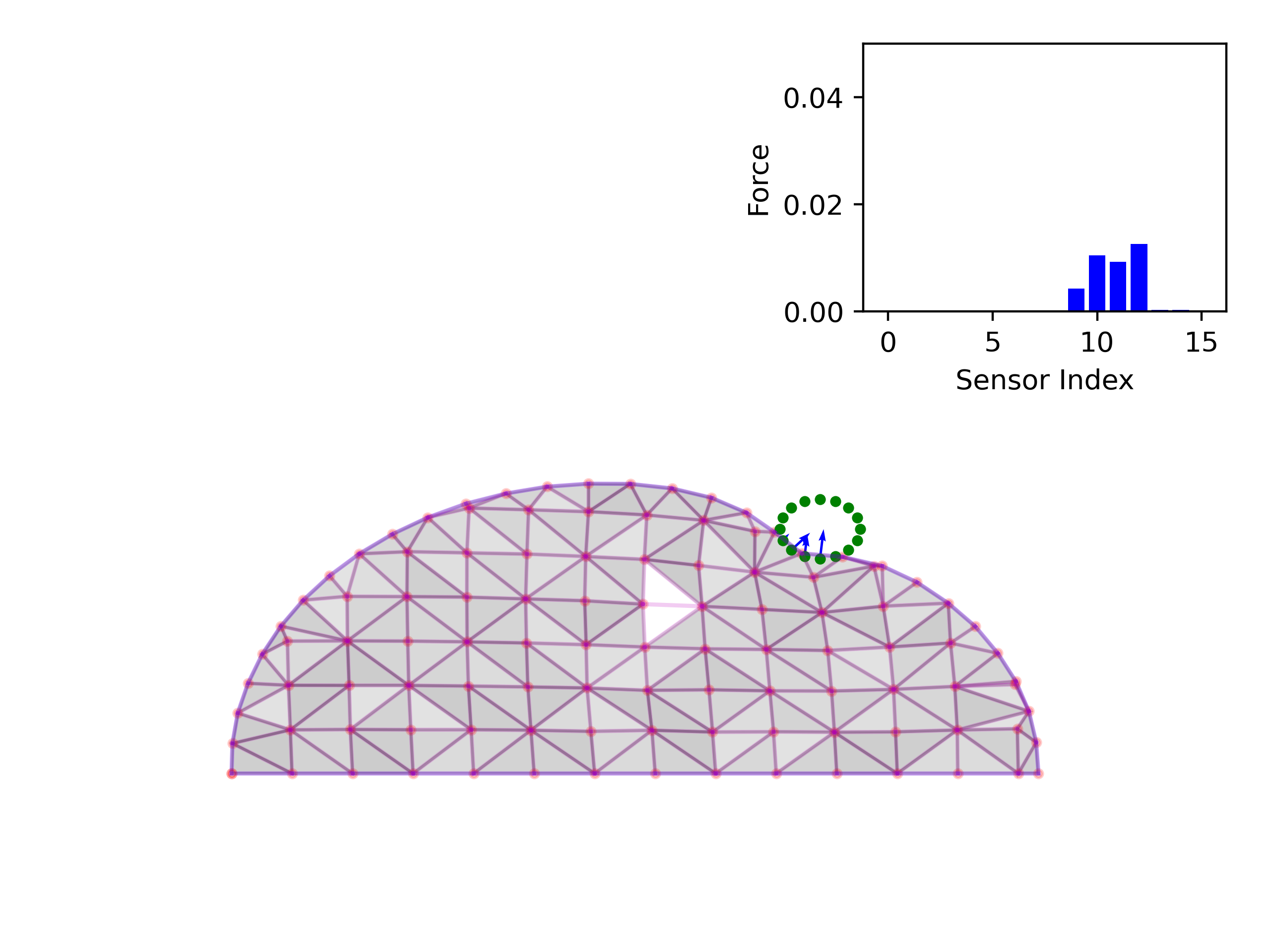}
    \end{subfigure}
    \begin{subfigure}[b]{0.19\textwidth}
        \includegraphics[width=\linewidth]{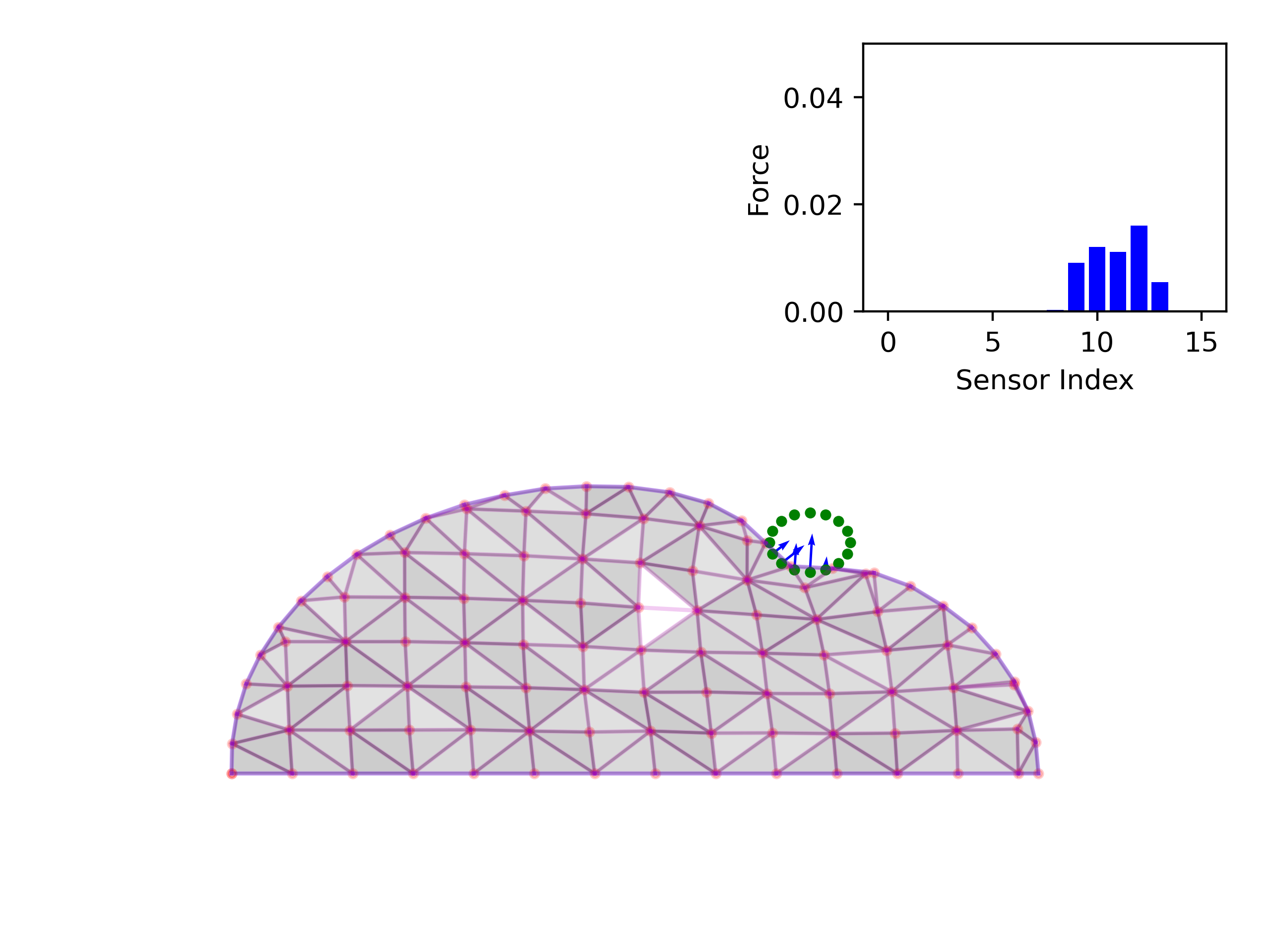}
    \end{subfigure}
    \caption{Comparison of selected simulation trajectories for big and small lumps. The simulation interface includes a force graph measured by the probe (including force vectors for easy visualization), the probe in green, and the breast model.}
    \label{fig:sim_vis_trajs}
\end{figure}

In \cref{fig:sim_vis_trajs} we visualize trajectories from two models in the simulation. As can be seen, the lump inside the breast model affects the sensed force, as expected. 

\subsection{Simulation Data Collection Hyperparameters} \label{app:simulation_data_collection}

The hyperparameters used for the simulation data collection are shown in \cref{tab:sim_params}.

\begin{table}[htbp]
\centering
\begin{subtable}[t]{0.45\textwidth}
\centering
\begin{tabular}{c|c}
\textbf{Name} & \textbf{Value} \\
\hline
$\sigma_{noise}$ & 0.0001 \\
$N_{points}$ & 16 \\
$R_{probe}$ & 0.1 \\
$\kappa_{collision}$ & 0.01
\end{tabular}
\caption{Probe hyperparameters}
\end{subtable}
\hfill
\begin{subtable}[t]{0.45\textwidth}
\centering
\begin{tabular}{c|c}
\textbf{Name} & \textbf{Value} \\
\hline
$\beta_1$ & 0.2 \\
$\beta_2$ & 0.999 \\
$lr$ & 0.001 
\end{tabular}
\caption{Adam optimizer hyperparameters}
\end{subtable}

\vspace{0.5cm}

\begin{subtable}[t]{0.45\textwidth}
\centering
\begin{tabular}{c|c}
\textbf{Name} & \textbf{Value} \\
\hline
$R_{Model}$ & $\mathcal{N}\left(1, 0.01\right)$ \\
$L_{grid}$ & 0.15 \\
$L_{perimeter}$ & 0.1 \\
$N_{points}$ & 0.001  \\
$\mu_{ym}$ & $U\left[0.0027, 0.0033\right]$ \\
$\sigma_{ym}$ & 0.0002 \\
$\mu_{pr}$ & $U\left[0.09, 0.11\right]$ \\
$\sigma_{pr}$ & 0.01
\end{tabular}
\caption{Breast-model hyperparameters}
\end{subtable}
\hfill
\begin{subtable}[t]{0.45\textwidth}
\centering
\begin{tabular}{c|c}
\textbf{Name} & \textbf{Value} \\
\hline
$ym_{lump}$ & 0.01 \\
$pr_{lump}$ & 0.1 \\
$p_{change}$ & 0.1 \\
$center_{lump}$ & $Ring\left(0.44, 0.55\right) $ \\
$R_{lump-change}$ & $U\left[0.07,0.15\right] $ \\
$R_{lump-no-change}$ & $U\left[0.11,0.21\right]$ \\
$\Delta_{R_{lump}}$ & $\mathcal{N}\left(0.03, 0.01\right)$
\end{tabular}
\caption{Lump hyperparameters}
\end{subtable}

\caption{Hyperparameters used for the simulation data collection}
\label{tab:sim_params}
\end{table}

\subsection{Simulation Imaging Results} \label{sec:sim_imaging_results}
\begin{figure}[htbp]
    \centering
    \vspace{-1em}
    \begin{subfigure}{0.38\textwidth}
        \includegraphics[width=\linewidth]{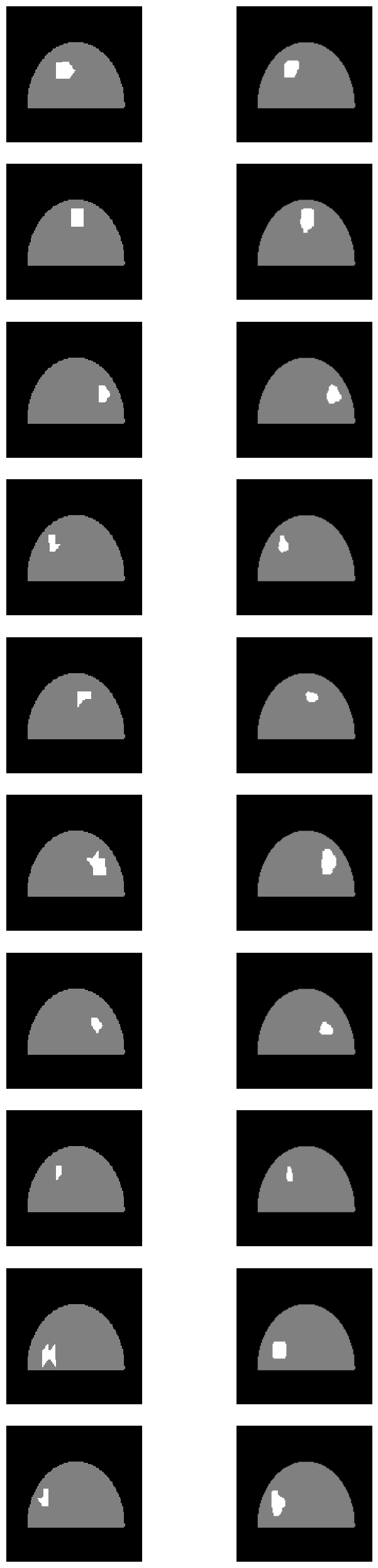}
    \end{subfigure}
    \hfill
    \begin{subfigure}{0.38\textwidth}
        \includegraphics[width=\linewidth]{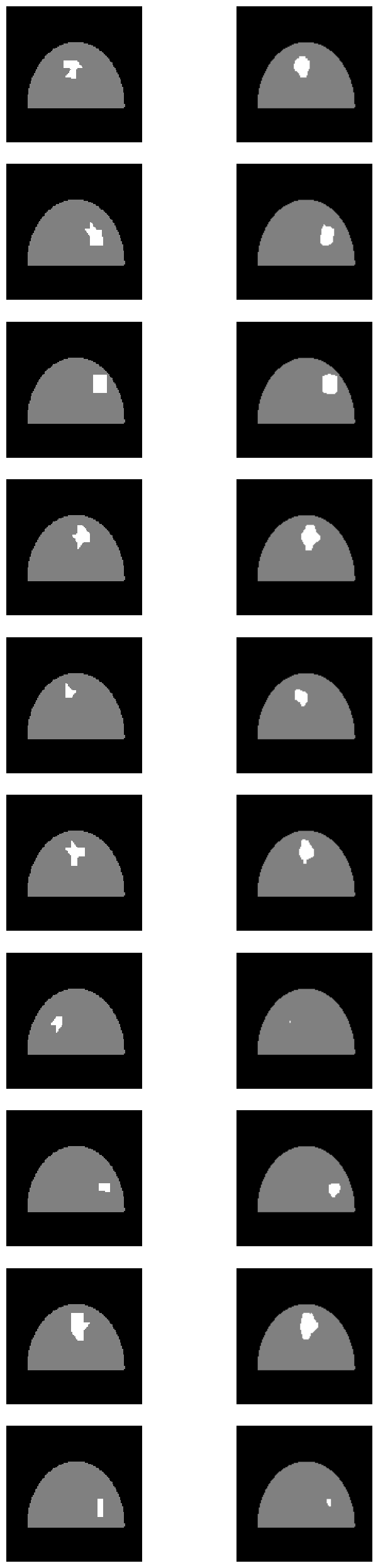}
    \end{subfigure}
    \caption{20 random test results of our simulation tactile imaging predictions. The ground-truth model image and our prediction are on the left and right of each column, respectively.}
    \label{fig:all_sim_imaging_results}
\end{figure}

To visualize the results shown in \cref{sec:results}, we present in \cref{fig:all_sim_imaging_results} 20 randomly sampled model-image reconstruction results in the simulation. As we can see our image predictor can reconstruct the model image fairly well from our tactile sequence representation. 

\subsection{Simulation Change Detection} \label{sec:sim_change_det_results}
\begin{figure}[htbp]
    \centering
    
    \includegraphics[width=0.98\textwidth]{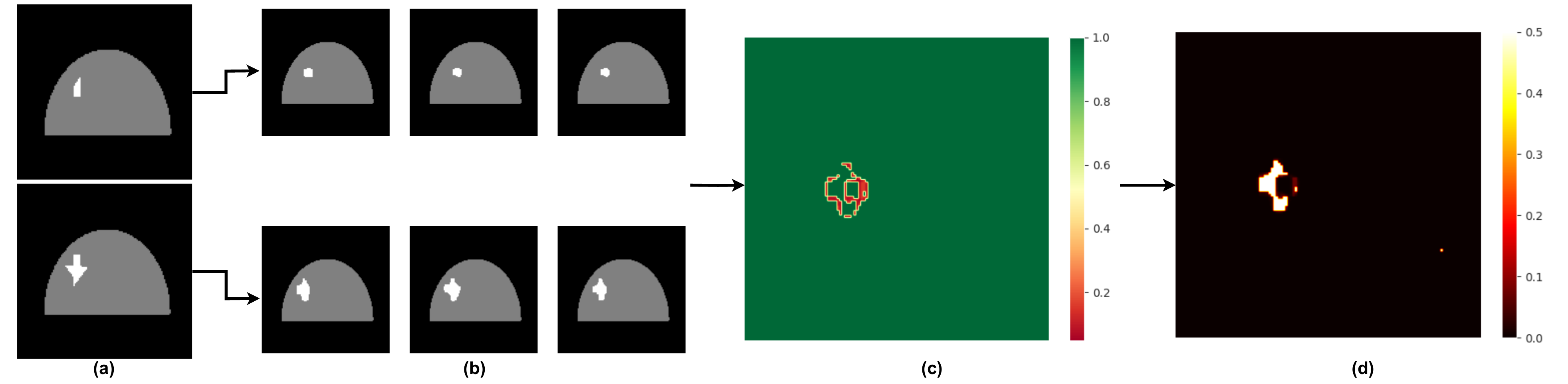}
    
    \caption{Our simulation change detection algorithm. (a) Two ground truth model images (b) Prediction of multiple images using the perturbations augmentation from each of the trials (c) Generation of a confidence map (d) Change map generation}
    \label{fig:change_det_sim}
\end{figure}

Since our image reconstruction in simulation accurately predicts the lump shape and location, we can use the image prediction for change detection and localization. As can be seen in \cref{fig:change_det_sim}, given two trials from the same model, we first generate multiple image predictions by taking a different permutation of the trial trajectories (we used 10 permutations in practice). Using the multiple prediction, we can generate a joint confidence map:

$$
c[i] = \frac{C}{C+\frac{1}{2}\left(\sigma_1[i]+\sigma_2[i]\right)}
$$

Where $c[i]$, $\sigma_1[i]$ and $\sigma_2[i]$ are the confidence map, first trial standard deviation and second trial standard deviation at pixel $i, \forall i\in \left[1,128 \cdot 128\right]$, respectively (the standard deviations are calculated across the different predicted image of each trial). The confidence calculation behaves such that very high and low standard deviations correspond to a confidence of 0 and 1, respectively, and when $\sigma_1[i]=\sigma_2[i]=C$ we get $c[i]=\frac{1}{2}$. Finally, each pixel gets a score:

$$
s[i] = |\mu_1[i]-\mu_2[i]|\cdot c[i]
$$

Where $\mu_1$ and $\mu_2$ are the average pixel value at pixel $i, \forall i\in \left[1,128 \cdot 128\right]$. 
Using the scores we can plot a change score map and localize changes. 
\begin{figure}[htbp]
    \centering
    
    \includegraphics[width=0.4\textwidth]{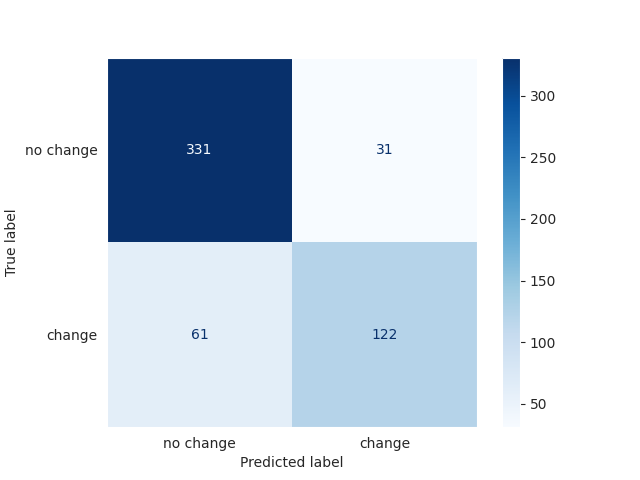}
    
    \caption{Our change prediction confusion matrix, when setting the threshold on the change score to 0.1}
    \label{fig:change_det_sim_results}
\end{figure}

To aggregate the change map to a single score for the two trials, we simply take the mean over all pixels. We report the confusion matrix when setting the threshold to $0.1$ in \cref{fig:change_det_sim_results}.

\section{Human Study Procedure and Results} \label{app:human_study}

\paragraph{Procedure}
The full procedure and given instructions for each participant are detailed below:

For explanation purposes, a single insert is put into a single shell.
\begin{tcolorbox}[colback=gray!10, colframe=gray!50, arc=4mm]
Our research is about detecting breast cancer using artificial palpation. We have breast phantoms, consisting of shells with inserts inside them, where each insert has a lump inside it. All lumps are spherical, and change in diameter and location. There is always a single lump inside of the insert. Your task will be to try and detect changes in the lump after a given amount certain amount of time. The lump can only grow larger or stay the same; it cannot change location, nor shrink. The lump will change with a $0.5$ probablity, otherwise it will stay the same. Please follow these instructions while palpating:

\begin{itemize}

\item Use only one finger when palpating

\item You can press however you like and for as long as you want to, but use reasonable force not to tear the phantom

\item In your answer, state “change” or “no change”

\end{itemize}

Here, there is a phantom with a lump inside, just as an example, it is not part of the trial, you can feel it to get a sense of the task.

\end{tcolorbox}

Next, the insert is changed, with the smallest possible change in the lump size (to calibrate the participants answers).
\begin{tcolorbox}
Now we have replaced the insert to have a larger lump size, such that the change in size is the lowest possible change.
You can now palpate to understand what a change might feel like.
\end{tcolorbox}

Finally, the following steps are repeated $\sim 4$ times:

\begin{enumerate}

\item A completely new insert is put into a phantom (with a random lump location, orientation, and size) 

\item The participant palpates the phantom

\item After the time interval, the lump, with 0.5 probability the insert is replaced with a different insert with the same location and orientation, but a larger size, and with 0.5 probability, the insert is taken out and put back in.

\item The participant palpates the phantom again, and their answer is logged together with the experiment metadata.  

\end{enumerate}

\paragraph{Results}

\begin{wraptable}{r}{0.4\textwidth}
\centering
\begin{tabular}{>{\centering\arraybackslash}m{1.5cm}|
                >{\centering\arraybackslash}m{1.5cm}|
                >{\centering\arraybackslash}m{1.5cm}}
 &  \textbf{Humans} &  \textbf{Ours} \\ \hline
\textbf{FAR} $\downarrow$ & $0.32$ & $0.19$ \\ \hline
\textbf{Recall} $\uparrow$ & $0.62$ & $0.82$ \\ 
\end{tabular}
\caption{Change detection recall and FAR for human participants and our approach.}
\label{tab:changedethuman}
\end{wraptable}
We had $16$ participants, and as explained above, each participant repeated the experiment up to $6$ times (with different time intervals). The results were 23/34 ($\sim 68\%$) correct non-changing classifications and 18/29 ($\sim 62\%$)correct changing classifications.

table

\paragraph{Ethical Concerns}
The human study was approved by the Technion's Institutional Review Board (IRB), approval number 2025-068. There are no dangers to participants. The experiment is equivalent to playing with a squishy toy for 
several minutes. The materials are safe – silicone, and the inner filling (which is not touched by 
participants) made out of gel that is commonly sold as a children’s toy (Borax, water, and glue). There
is absolutely no inconvenience of any kind to the participants during the experiment.
This experiment allows us to calibrate the performance of our system in comparison to 
human skills, which is important for understanding the potential of our research. Each participant received financial compensation above the minimum wage. The data recorded is kept private on a secure server, which only contains the names and emails of
the participants and answers to questions. All files are password-protected, with passwords only 
known to the relevant researchers. The results we publish are summarized statistics, and do not 
contain any personal information about participants. Each participant interacted directly with one of
the researchers in the project. We explained verbally and in written instructions to participants that their participation is voluntary 
and that they can withdraw at any time with no negative consequences for them.

\section{Self-supervised Training Technical Details} \label{sec:selfsup_details}
\begin{figure}[htbp]
    \centering
    
    \includegraphics[width=0.5\textwidth, trim=20 0 10 0, clip]{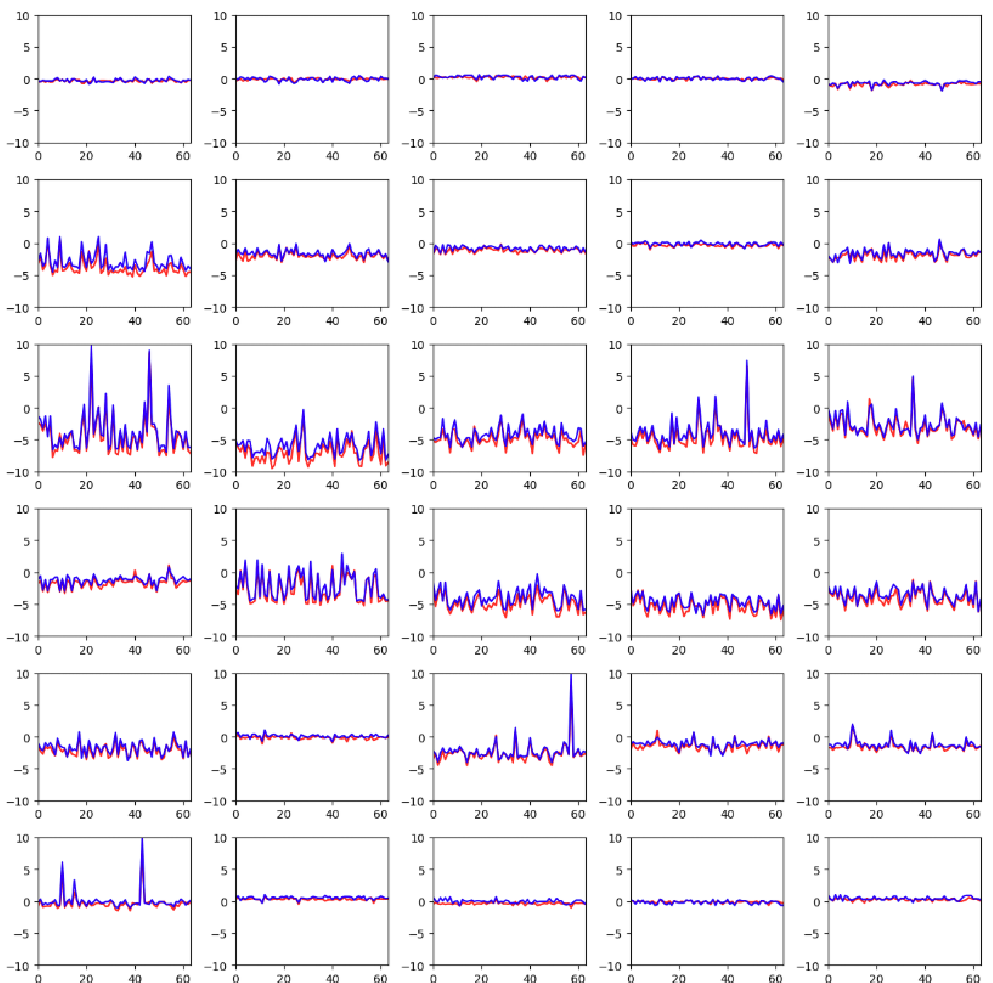}
    
    \caption{Force reconstruction on real data of the $z$-axis component of each of the $30$ sensors from the representation in the last time step ($z_N$). In blue and red are the true and reconstructed forces, respectively.}
    \label{fig:force_recon_last}
\end{figure}

Our self-supervised architecture, as shown in \cref{fig:method_rep}, is composed of the FLE (Force Location Encoder), a sequence encoder, and a force decoder.
The FLE is simply adding a linear projection for the forces and a basic sinusoidal Positional Encoding (PE) for the $6$-dim locations (both the projection and the PE are of size $256$). The sequence encoder is a one layer RNN with a GRU and with a $1024$ hidden size. 
The force decoder first linearly projects the input representation and uses PE on the desired reconstruction to size $1020$, and adds both together. Next, a three-layer MLP is used (with a $2048$, $1024$ hidden sizes) to predict the forces at the desired location. An MSE loss (with random indices sampling) is used as shown in \cref{sec:method_rep}. 

A visualization of the results of the reconstruction from the last representation is shown in \cref{fig:force_recon_last}.

\section{Downstream Tasks Technical Details} \label{sec:downstream_tasks_details}
To reconstruct the MRI images, we use the pre-trained (and frozen) encoder from the self-supervised step. We aim to learn a mapping from the last-step representation $z_T$ to the model MRI image.
To do so, we use a conditional flow matching-inspired architecture to reconstruct the $128\times 128$ image from the vector representation. 
During training, we sample a normally distributed $128\times 128\times 1$ noise image, and recursively pass it through a unet \citet{unet} to map it to an $128\times 128\times L$ latent image. Finally the image logits are calculated using a small CNN decoder. The network is trained using a cross-entropy loss.  
After training, in order to sample an image, we simply draw a single noise sample and map it using the learned conditional mapping to an image.
The full forward pass of our proposed predictor is in \cref{alg:forward_pass}
For the tactile imaging problem, we randomly split the trials to train and test (with the same split for the representation learning and the imaging). As for the change detection, we split per insert configuration, where all trials from the same configuration are either in train or test (this is done to have negative examples in the change detection task).

\begin{algorithm}
\caption{Forward pass of our image prediction.}
\label{alg:forward_pass}
\begin{algorithmic}[1]

\State \textbf{Input:} Representation tensor $\mathbf{r} \in \mathbb{R}^{B \times D}$
\State \textbf{Output:} Image logits $\mathbf{I} \in \mathbb{R}^{B \times C \times H \times W}$

\State $\mathbf{z} \gets \text{sample\_noise}(B, Z, H, W)$ \Comment{Sample Gaussian noise: $B \times Z \times H \times W$}
\State $\text{time\_steps} \gets \text{linspace}(1, 0, N_t)$

\For{each $t$ in $\text{time\_steps}$}
    \State $t \gets t$ expanded to shape $B \times 1 \times H \times W$ 
    \State $\mathbf{z} \gets \mathbf{z} + \text{flow\_predictor}(\text{concat}[\mathbf{z}, t], \mathbf{r})$ 
        \Comment{UNet conditioned on time and representation}
\EndFor

\State $\mathbf{I} \gets \text{decoder}(\mathbf{z})$ \Comment{Decoder: Conv2d $\rightarrow$ ReLU $\rightarrow$ Conv2d to logits}
\State \textbf{return} $\mathbf{I}$

\end{algorithmic}
\end{algorithm}

\section{Architecture Ablation}
To motivate the chosen architectures, we perform an ablation study, both for the representation learning and for the tactile imaging.
For the representation learning architecture we tested both an RNN with GRU (\textbf{GRU}) and a two-layer vanilla transformer (\textbf{TR}). For both architectures, we also considered a version where some of the input poke trajectories (e.g 20\%) are masked, and we reconstruct only these. 
For the tactile imaging architecture, we considered the flow matching (\textbf{FM}) and a series of transposed convolutions (\textbf{TC}).
We perform the training for the ablation study using $\times 0.5$ of our full training data due to cost and time concerns.

The results of the ablation study can be seen in \cref{tab:arch_ablation}. While this ablation study is preliminary, we chose the GRU and flow matching models for the representation learning and tactile imaging, respectively. We leave further investigation into more advanced architecture to future work.   

\begin{table}[h]
\centering
\caption{Lump Size and Center-of-Mass (CoM) errors. We report standard deviation of the sample mean across 3 random seeds.}
\vspace{0.3cm}
\label{tab:arch_ablation}
\begin{tabular}{lcc}
\toprule
\textbf{Method} & \textbf{Size Error [\%] $\downarrow$} & \textbf{CoM Error [mm] $\downarrow$}  \\ 
\midrule
GRU + FM (Ours) & $20.9 \pm 0.9$ & $3.6 \pm 0.1$ \\
GRU + TC & $27.9 \pm 1.8$ & $3.5 \pm 0.1$  \\
TR + FM & $35.4 \pm 0.7 $ & $6.7 \pm 0.3$  \\
Masked TR + FM & $57.9 \pm 2.0$ & $9.9 \pm 0.1$ \\
Masked GRU + FM & $23.9 \pm 1.0 $ & $4.0 \pm 0.1 $ \\
\hline
\end{tabular}
\end{table}

\section{Tactile Imaging Results} \label{sec:imaging_results}
\begin{figure}[htbp]
    \centering
    \vspace{-1em}
    \includegraphics[width=1\textwidth]{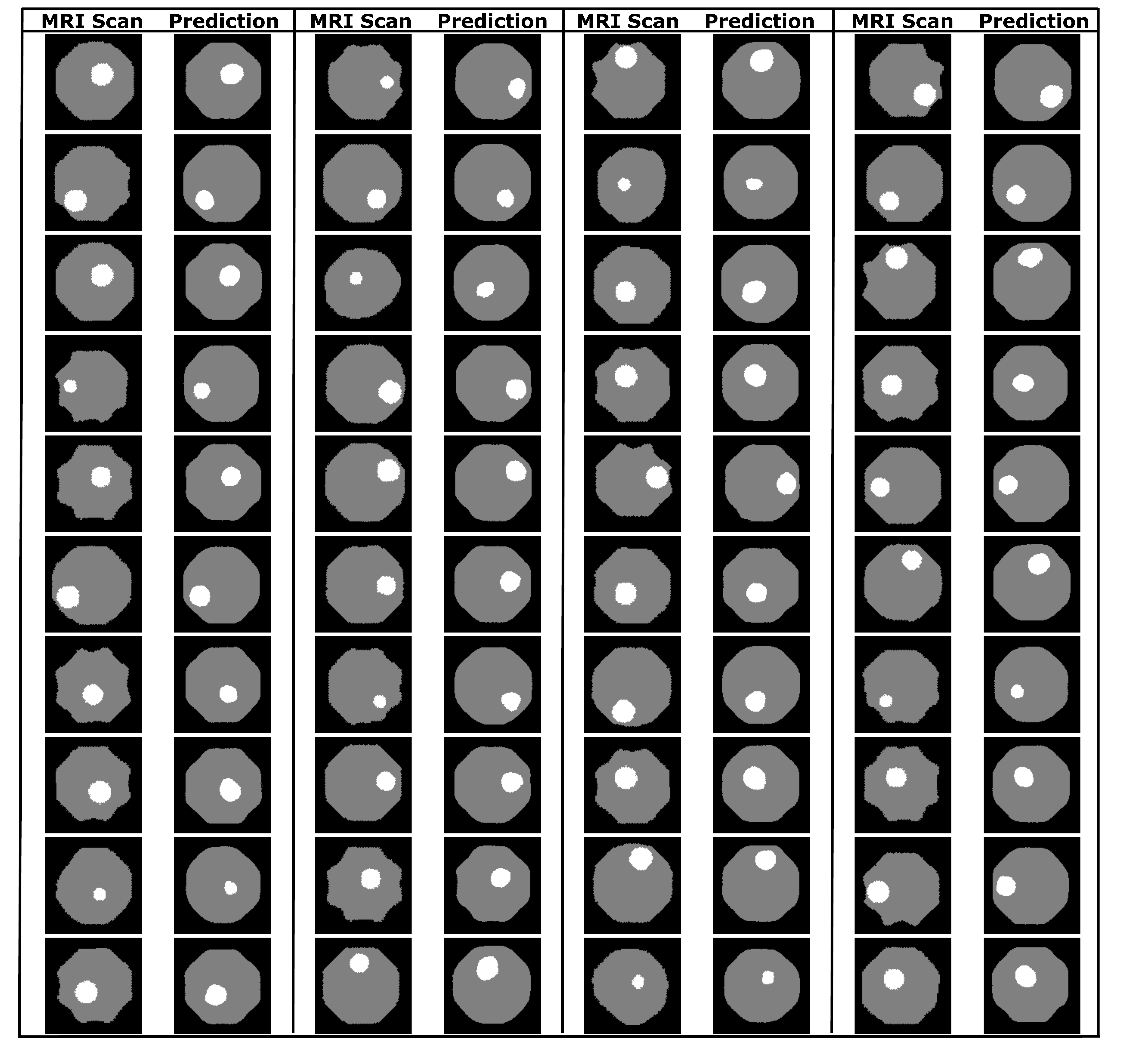}
    \caption{40 random test results of our MRI predictions. The ground-truth MRI image and our prediction are on the left and right of each column, respectively.}
    \label{fig:all_imaging_results}
\end{figure}

To visualize the results shown in \cref{tab:image_prediction}, we present in \cref{fig:all_imaging_results} 40 randomly sampled MRI prediction results.
The predictions are accurate, in terms of lump location and size, as was quantitatively shown in \cref{tab:image_prediction}.

\section{Force Map Generation Procedure} \label{sec:force_map_baseline}

\begin{figure}[H]
    \centering
    \includegraphics[width=\linewidth]{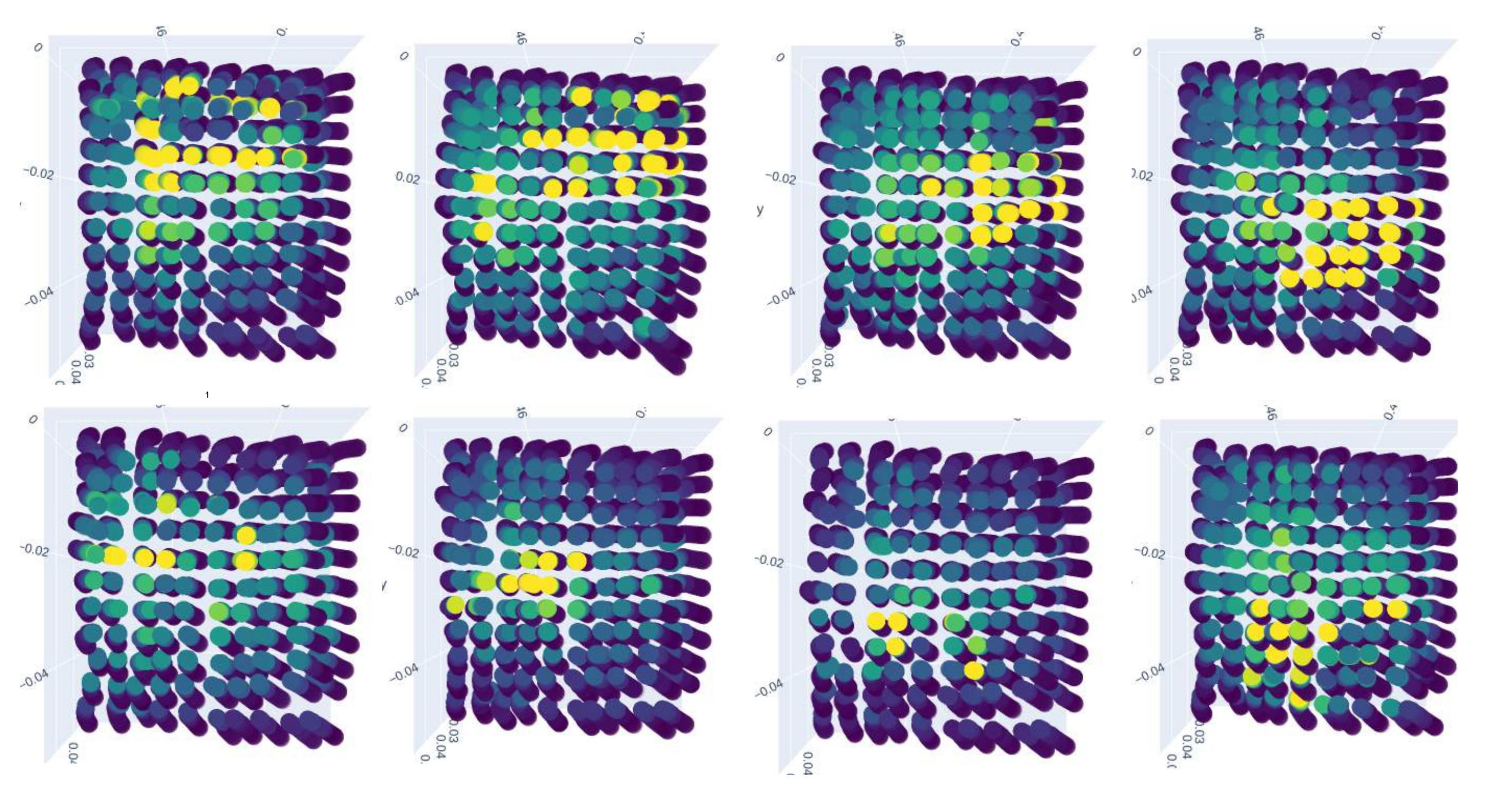}
    \caption{Top view of our 3D tactile visualization for a constant insert. Each visualization corresponds to a trial taken from the insert at a different orientation (from rotation $1$ to $8$, clockwise, starting from $1$ in the top left corner).}
    \label{fig:3d_tac_vis}
\end{figure}

The fact that tactile measurements are not interpretable in their raw form, makes it difficult to debug the data collection process. To this end we created a visualization tool for tactile data, collected from a single trial (with multiple trajectories).
We aggregate the tactile measurements at each location by taking the maximal $z$ force measured by all $30$ sensors. The result can be seen in \cref{fig:3d_tac_vis}, and an interactive 3D version can be found in the supplementary materials. 

\begin{figure}[H]
    \centering
    \vspace{-1em}
    \begin{subfigure}{0.32\textwidth}
        \includegraphics[width=\linewidth]{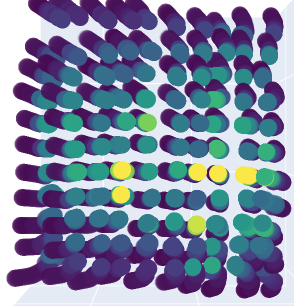}
        \caption{}
    \end{subfigure}
    \hfill
    \begin{subfigure}{0.32\textwidth}
        \includegraphics[width=\linewidth]{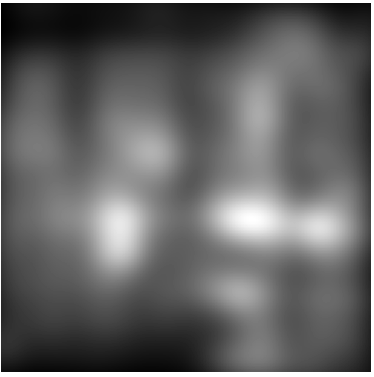}
        \caption{}
    \end{subfigure}
    \hfill
    \begin{subfigure}{0.32\textwidth}
        \includegraphics[width=\linewidth]{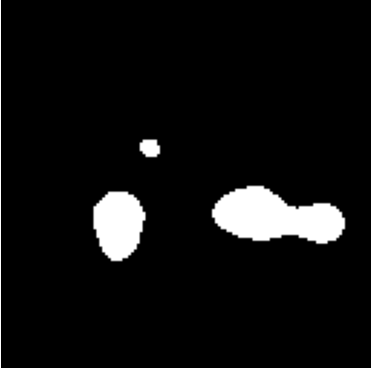}
        \caption{}
    \end{subfigure}
    \caption{Force map generation process. (a) The 3D forces visualization. (b) The KDE image. (c) The resulting image after taking a threshold over the KDE image.}
    \label{fig:binary_force_map}
\end{figure}

As can be seen in \cref{fig:3d_tac_vis}, the generated force map is correlated to the lump location. Hence, we hypothesized that it can be used as a baseline for the tactile-imaging problem. As a first simple baseline, we take the mean over the forces and locations of the last $10\%$ data points of each trajectory in the 3D map. Next, we use Kernel Density Estimation (KDE) to produce an image from individual measurements. Finally, we take a threshold to binarize the image (which we set to maximize test accuracy). The result can be seen in \cref{fig:binary_force_map}. As can be seen from the image, although the force map is very informative, it is not enough for accurate lump size and CoM prediction.

The result above illustrates that we cannot \textit{directly} use the force map to predict the lump properties. To further motivate our approach, we also show that the force map is an \textit{insufficient representation} for tactile imaging. To do so, we trained an identical flow matching predictor to the one we used on top of our learned representation, but with a flattened force map. The results can be seen in \ref{tab:image_prediction}. Clearly, the (non-learnable) force-map representation underperforms compared to our approach. We have also tried alternative architectures for the prediction, including a convolutional architecture on top of a non-flatten force-map representation.

\section{MRI Image Preprocessing} \label{sec:mri_preprocessing}

As explained in the main text, our imaging reconstruction is a classification problem, aiming to reconstruct one of the classes - background/body/lump for each pixel.
So, in order to use the MRI images as ground-truth, we first had to build a pre-processing tool. 

\begin{figure}[H]
    \centering
    \includegraphics[width=0.8\textwidth]{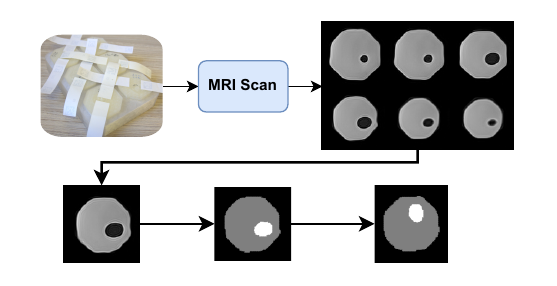}
    
    \caption{MRI pre-processing procedure. We 3D printed a hub, which can hold up to 6 inserts at a time for the MRI scan in order to save time and resources. After the MRI scan, we take a slice at a constant height and use Otsu's threshold \citet{otsu1975threshold} to binarize the image. Finally, we can rotate the scan to achieve any desired insert orientation. }
    \label{fig:mri_preprocess}
\end{figure}

\begin{figure}[htbp]
    \centering
    \includegraphics[width=0.8\textwidth]{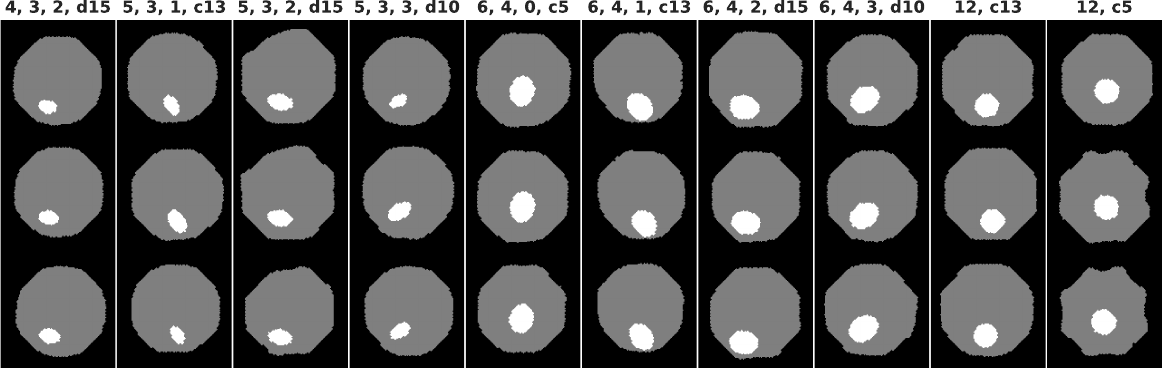}
    
    \vspace{1em} 
    
    \includegraphics[width=0.8\textwidth]{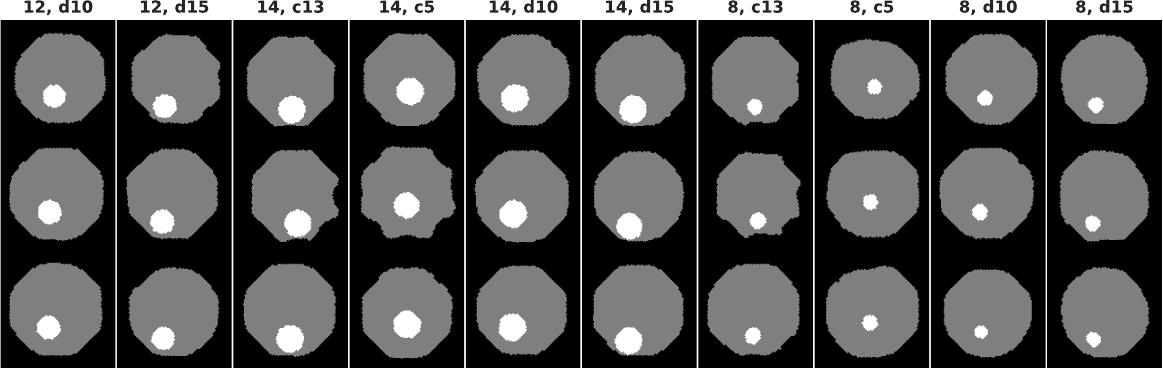}
    
    \caption{MRI images used as ground-truth, after pre-processing. Each insert was scanned 3 times. Above each trio, we show the first and second radii, the orientation, and location (either diagonal of center and radius from center) of the ellipsoid lump, where, in the case of a spherical lump, only a signed radius and orientations are shown.}
    \label{fig:mri_grid_outputs}
\end{figure}

Although MRI data is 3D in nature, since our lumps are manufactured with similar heights, we chose to work with a single horizontal slice for simplicity. The procedure for our MRI pre-processing is shown in \cref{fig:mri_preprocess}, and examples of the final result are shown in \cref{fig:mri_grid_outputs}

\section{Shell Classification}
\begin{figure}[htbp]
    \centering
    \includegraphics[width=0.5\textwidth]{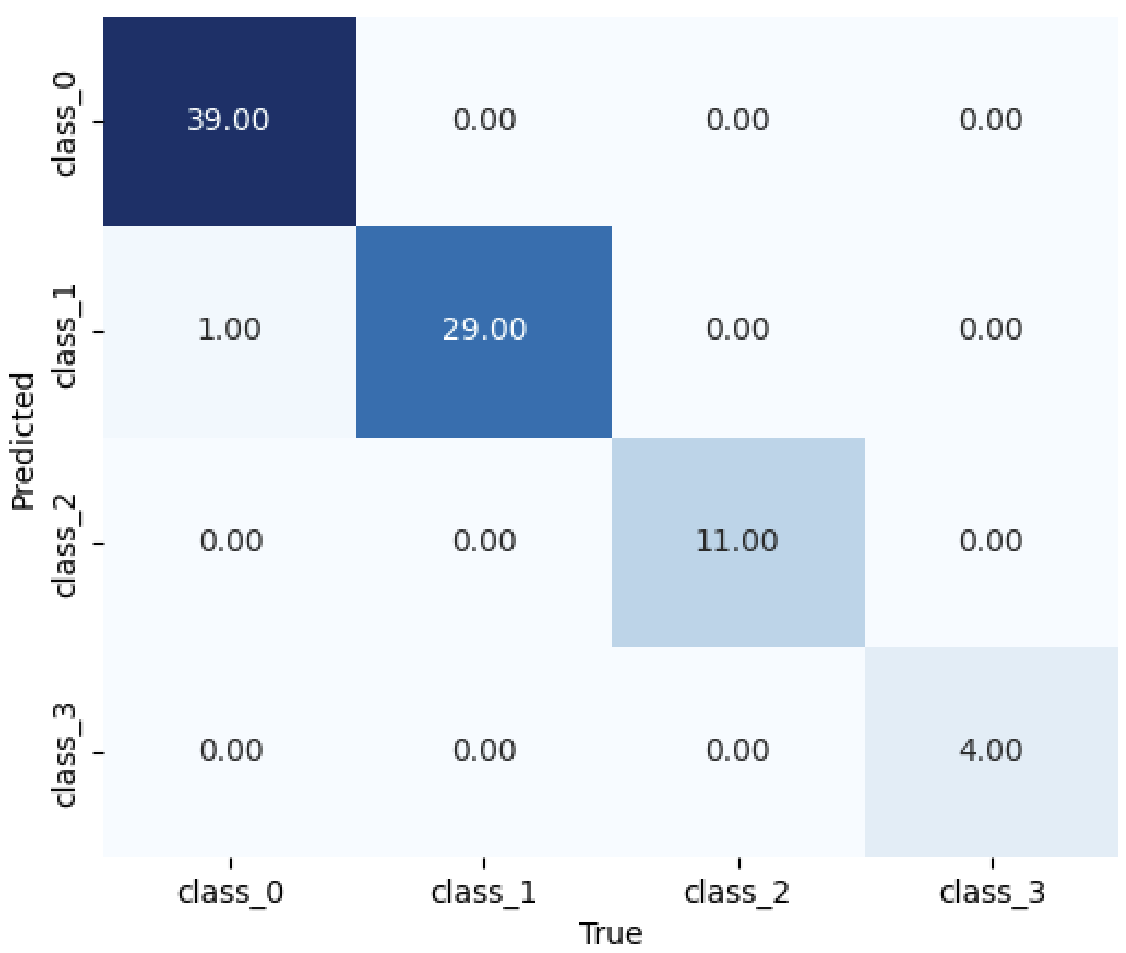}
    \caption{Confusion matrix for the shell classification downstream task on test sequences. Other than one mistake, the model has successfully classified all sequences.}
    \label{fig:shell_class}
\end{figure}

To further show the expressiveness of our extracted representation, we chose another downstream task that cannot be inferred from the tactile imaging output.
We aimed to classify the shell based on the tactile sequence.
We have trained a small MLP on top of the frozen representation with a cross-entropy loss, aiming to classify between the $4$ possible shells. 
Although the shells were manufactured in the exact same way, the representation is expressive enough to capture small manufacturing artifacts, effectively distinguishing between the shells with an accuracy of $99.6 \pm 0.7$ on test sequences.
The confusion matrix can be seen in \cref{fig:shell_class}
\label{sec:shell_classification}.

\section{Compute} \label{sec:compute}
The PalpationSim simulator runs on CPU. We collected data from it by running the simulation on $\sim 500$ CPUs to parallelize the process, but a single instance is very light-weight. 
All of the training procedures, including the self-supervised phase and the image reconstruction training ran on a cluster of 12 A4000 GPUs, although each single run needed only a single GPU to run.  